\documentclass[letterpaper]{article}

\usepackage{cite, graphicx, amsmath, amsfonts, amssymb, array, latexsym, algorithm, algorithmic}
\usepackage{multirow, dsfont, yfonts}

\newcommand{\TVi}{{\rm TV}_i}
\newcommand{\TVa}{{\rm TV}_a}
\newcommand{\TVL}{{\rm TV}_L}
\newcommand{\Hh}{\mathcal{H}}
\newcommand{\Aa}{\mathcal{A}}
\newcommand{\Rr}{\mathcal{R}}
\newcommand{\Uu}{\mathcal{U}}
\newcommand{\VV}{\mathbb{V}}
\newcommand{\UU}{\mathbb{U}}
\newcommand{\fft}{\mathcal{DFT}}
\newcommand{\UL}{\mathcal{U}_L}
\newcommand{\Rnm}{\mathbb{R}^{N \times M}}
\newcommand{\RR}{(\mathbb{R}^{N \times M})^2}
\newcommand{\RRL}{(\mathbb{R}^{N \times M})^{2 L}}
\newcommand{\argmin}{\operatornamewithlimits{argmin}}

\newcommand{\DIV}{{\rm div}_{\bar{\theta}_L}}
\newcommand{\GRD}{\nabla_{\bar{\theta}_L}}

\begin{document}

\title{An Iterative Shrinkage Approach to Total-Variation Image Restoration}

\author{Oleg~Michailovich
\thanks{This research was supported by a Discovery grant from NSERC -- The Natural Sciences and Engineering Research Council of Canada. Information on various NSERC activities and programs can be obtained from {\tt http://www.nserc.ca}.}%
\thanks{O. Michailovich is with the School of Electrical and Computer Engineering, University of Waterloo, Canada N2L 3G1 (phone: 519-888-4567; e-mails: olegm@uwaterloo.ca).}}

\maketitle

\begin{abstract}
The problem of restoration of digital images from their degraded measurements plays a central role in a multitude of practically important applications. A particularly challenging instance of this problem occurs in the case when the degradation phenomenon is modeled by an ill-conditioned operator. In such a situation, the presence of noise makes it impossible to recover a valuable approximation of the image of interest without using some a priori information about its properties. Such a priori information -- commonly referred to as simply {\em priors} -- is essential for image restoration, rendering it stable and robust to noise. Moreover, using the priors makes the recovered images exhibit some plausible features of their original counterpart. Particularly, if the original image is known to be a piecewise smooth function, one of the standard priors used in this case is defined by the Rudin-Osher-Fatemi model, which results in total variation (TV) based image restoration. The current arsenal of algorithms for TV-based image restoration is vast. In the present paper, a different approach to the solution of the problem is proposed based on the method of iterative shrinkage (aka iterated thresholding). In the proposed method, the TV-based image restoration is performed through a recursive application of two simple procedures, {\it viz.} linear filtering and soft thresholding. Therefore, the method can be identified as belonging to the group of first-order algorithms which are efficient in dealing with images of relatively large sizes. Another valuable feature of the proposed method consists in its working directly with the TV functional, rather then with its smoothed versions. Moreover, the method provides a single solution for both isotropic and anisotropic definitions of the TV functional, thereby establishing a useful connection between the two formulae. Finally, it is shown experimentally that, in the case when image degradation is caused by blur, the proposed method can provide restoration results of superior quality as compared to the case of sparse-wavelet deconvolution.
\end{abstract}


\section{Introduction}
Both environmental effects and imperfections of image acquisition devices tend to degrade the quality of imagery data, thereby making the problem of image restoration an integral part of modern imaging sciences. In particular, medical imaging~\cite{Campisi07}, astronomical~\cite{Vogel02} and laser~\cite{Roggeman96} imaging, microscopy~\cite{Sluder07}, remote sensing~\cite{Twomey02}, and photography~\cite{Ben-Ezra04} are all examples of applications in which the necessity to improve the resolution and contrast of digital images routinely arises. Despite the relative ``antiquity" of the theory of image restoration (with the first papers on the subject having been published as far back as at the end of the 60s~\cite{Morgan67, Rino69}), there still exists a need for its further advancement via proposing new approaches as well as by improving the computational efficiency of existing ones. Addressing the second of the above two objectives forms the core of the developments presented here.

The algorithm reported in this manuscript is based on a standard linear measurement model, in which the original image $f$ and its measurements $g$ are considered to be elements of a (finite dimensional) signal space $\UU$, and they are assumed to be related according to 
\begin{equation}\label{mdl}
g = \mathcal{H}\{f\} + e, 
\end{equation}
where $\mathcal{H}: \UU \rightarrow \UU$ is a bounded operator describing the effect of image degradation and $e$ stands for both modeling and measurement noises. The problem of recovering $f$ from $g$ becomes particularly challenging in the case when the operator $\mathcal{H}$ is either ill-conditioned or singular, in which case the problem of recovering $f$ is commonly referred to as {\em ill-posed}~\cite{Hansen98}. On such conditions, a standard way to proceed with the solution of (\ref{mdl}) is to recover $f$ {\em approximately} by solving the following variational problem
\begin{equation}\label{regul}
f \simeq \argmin_{u \in \UU} \left\{ \frac{1}{2}\left\| \Hh\{u\} - g \right\|_2^2 + \lambda \, \varphi(u) \right\} 
\end{equation}
where $\| \cdot \|_2$ denotes the $\ell_2$-norm and $\varphi: \UU \rightarrow \mathbb{R}$ is a convex lower semicontinuous function on $\UU$ whose role is to render the solution of (\ref{regul}) unique and stable. Moreover, minimizing $\varphi$ has the effect of restricting the solution of (\ref{regul}) to functions of a predefined class which the original image $f$ is believed to belong to. In this case, the parameter $\lambda>0$ (which is conventionally referred to as a {\em regularization parameter}) controls the balance between the model- and prior-dependent terms in (\ref{regul}).   

Probably the most renowned definition of $\varphi$ as an $\ell_2$-type norm dates back to the works of A. N. Tikhonov~\cite{Tikhonov63}. In application to image processing, however, this choice is rarely used in current practice because of the property of resulting solutions to have overly smoothed edges. In this regard, a more successful choice of $\varphi$ would be the one that allowed recovering $f$ while maximally preserving its fine details. Following this line of considerations, Rudin {\it et al}~\cite{Rudin92} proposed to define the regularization functional $\varphi$ to be the {\em total variation} (TV) seminorm~\cite{Ziemer89}, {\it viz.}
\begin{equation}\label{iTV}
\varphi(u) \equiv \TVi(u) := \Big\| \sqrt{|u_x|^2 + |u_y|^2} \Big\|_1,
\end{equation}
where $u_x$ and $u_y$ denote the partial derivatives of $u \in \UU$ and $\| \cdot \|_1$ stands for the $\ell_1$-norm. The resulting model -- commonly referred to as the Rudin-Osher-Fatemi (ROF) model -- is nowadays considered to be one of the most fundamental models of modern imaging sciences. Along with its {\em isotropic} version~\cite{Esedoglu04}, i.e.
\begin{equation}\label{aTV}
\varphi(u) \equiv \TVa(u) := \big\| |u_x| + |u_y| \big\|_1,
\end{equation}
the ROF model has proven to be an extremely useful tool in numerous applications such as image de-blurring~\cite{Vogel02}, ``$u+v$" decomposition~\cite{Meyer01}, super-resolution~\cite{Farsiu04}, and image impainting~\cite{Bertalmio00}, just to name a few.

Minimization of the TV-regularized functional $E(f) = \big\{ \frac{1}{2}\left\| \Hh\{f\} - g \right\|_2^2 + $ $\lambda \, {\rm TV}(f) \big\}$ (with ${\rm TV} \in \{ \TVi, \TVa \}$) is known to be a relatively difficult optimization problem because of the non-differentiability of the TV regularizer. In order to overcome this difficulty, some methods substitute the absolute value function in (\ref{iTV}) and (\ref{aTV}) by its smooth approximation~\cite[Section 8.2]{Vogel02},~\cite{He05, Rodriguez09}. Even though using such approximations provides an access to a variety of efficient tools of smooth optimization, in order to perform stably, the resulting computational schemes require the use of proper preconditioning procedures, which makes these methods ``costly" for fast processing of standard-size images (e.g. $256\times 256$ or $512\times 512$). The same concern can be extended to the algorithms employing the tools of constrained optimization~\cite{Chan96, Goldfarb06, Goldfarb09}.

Practical problems related to the size of imagery data have motivated the community of imaging scientists to reconsider the potential of some {\em first-order} image reconstruction algorithms. In particular, the methods detailed in~\cite{Chambolle97, Osher05, Darbon06, Aujol09} are capable of finding a solution to (\ref{regul}) by means of simple recursive procedures, which make them particularly attractive for processing of large amounts of data. Unfortunately, these methods are only applicable to the de-noising setting (i.e. $\Hh$ is an identity), with their extension to the case of non-trivial, rank-deficient $\Hh$ being currently considered impossible~\cite{Aujol09}. 

A different approach to the solution of (\ref{regul}) was recently proposed in~\cite{Bioucas-Dias06} based on the majorization-minimization (MM) method~\cite{Hunter04}. In this case, a direct minimization of $E(f)$ is substituted by recursively minimizing its quadratic majorizer whose minima can be found via solution of a system of linear equations. For typical size images, however, the system can only be solved iteratively (using, e.g., the conjugate gradient algorithm), which substantially increases the overall computational cost of the procedure. It is also interesting to note that, even though derived from a different perspective, the method of~\cite{Bioucas-Dias06} is essentially identical to the method of lagged-diffusivity~\cite{Vogel96, Vogel02}. Furthermore, a stable implementation of this method requires using a smoothed version of the TV functional (the fact not mentioned in~\cite{Bioucas-Dias06}), which is necessary to prevent the diffusivity coefficients from becoming unbounded.

It is important to note that the recent interest in application of MM-type strategies to image restoration seems to have been triggered by the works reported in~\cite{Daubechies03, Figueiredo03} (see also~\cite{Elad07} for a nice summary of the subject). In these works, the image restoration is performed under the assumption that $f$ can be {\em sparsely} represented in the domain of a certain linear transform, which leads to the definition of $\varphi$ in (\ref{regul}) as the $\ell_1$-norm. The most remarkable result of these studies has been in showing that using the MM method allows one to solve the above problem by means of a simple first-order procedure. The latter -- known as {\em iterated shrinkage} (aka {\em iterative thresholding}~\cite{Blumensath08}) -- consists of repetitive application of two simple steps: a back-projection correction and soft thresholding. Moreover, under a few standard assumptions, the iterated shrinkage is guaranteed to converge to a minimizer of the $\ell_1$-constrained cost functional.     

Despite the conceptual similarity between the TV- and $\ell_1$-norm based regularizers~\cite{Steidl04}, an iterative shrinkage approach to minimization of the cost functional in (\ref{regul}) still seems to be missing. Thus, the main question addressed in the present study is whether or not it is possible to solve (\ref{regul}) for the case of non-trivial $\Hh$ by means of an iterative shrinkage (IS) scheme. As will be shown below, in the discrete setting, the above question can be answered affirmatively. Accordingly, introducing an IS scheme for TV-based image restoration forms the main contribution of this work. Moreover, the proposed algorithm can be used to solve (\ref{regul}) for the cases of {\em both} $\varphi=\TVi$ and $\varphi=\TVa$. In fact, the proposed numerical scheme will include a single scalar parameter which allows a transition between the isotropic and anisotropic cases. Hence, another contribution of this paper consists in demonstrating a connection between $\TVi$- and $\TVa$-regularizers. Finally, it will be proven conceptually and experimentally that, in the case of ill-conditioned $\Hh$, the TV-based restoration can be expected to provide better reconstruction results as compared to the case of sparse wavelet regularization~\cite{Figueiredo03, Elad07}.   

The remainder of the paper is organized as follows. Section II provides a number of essential technical details which are necessary for the developments in subsequent sections of the paper. An iterative shrinkage algorithm for the solution of (\ref{regul}) with $\varphi=\TVi$ is detailed in Section III, whereas Section IV extends these results to the case of isotropic TV. Some important details regarding the implementation of the proposed method are discussed in Section V. Finally, the results of comparative experiments are summarized in Section VI, while Section VII finalizes the paper with a discussion and conclusions. 

\section{Technical Preliminaries}
\subsection{Signal Space}
In most of the practically important settings, images are finite dimensional objects. For this reason, we formulate our approach under the assumption that both original and measured images belong to the vector space of real-valued $N\times M$ matrices. Moreover, we endow this space (referred below to as $\UU$) with the standard inner product $\langle f, g\rangle = \sum_{n=0}^{N-1}\sum_{m=0}^{M-1} f_{n,m} \, g_{n,m}$ and require that all the vectors in $\UU$ are bounded and hence possess a finite $\ell_2$-norm, defined in the standard way as $\|f\|_2 = \sqrt{\langle f, f \rangle}$. Finally, the elements of the signal space are also constrained to have {\em zero} mean value, which leads to a formal definition of $\UU$ as
\begin{equation}\label{space}
\UU = \left\{ f \in \Rnm \mid \left\langle \mathbf{1}, f \right\rangle = 0, \, \|f\|_2<\infty \right\},
\end{equation}
where $\bf 1$ denotes an $N\times M$ matrix of ones. We note that the assumption of zero mean should not be regarded as a restrictive one, since in practice it is rarely a problem to subtract the mean value from a data image, as well as to re-normalize a zero-mean image to make its values saturate a required range, e.g., $[0, 255]$.

Let $f_x$ and $f_y$ be the partial differences of $f$ taken in the column and row direction, respectively. Then, the discrete versions of the TV functionals (\ref{iTV}) and (\ref{aTV}) can be defined as
\begin{equation}\label{iTVd}
\TVi(f) = \Big\langle \mathbf{1}, \sqrt{|f_x|^2 + |f_y|^2} \Big\rangle,
\end{equation}
and 
\begin{equation}\label{iTVd}
\TVa(f) = \big\langle \mathbf{1}, |f_x| + |f_y| \big\rangle.
\end{equation}
Consequently, the problem of TV-based image restoration of $f$ in (\ref{mdl}) can be restated as computing
\begin{align}\label{TVgen}
f_{\rm TV} &= \argmin_{f \in \UU} \left\{ E(f) \right\}, \mbox { where } \notag \\
E(f) = & \frac{1}{2}  \left\| \Hh \{f\} - g \right\|_2^2 + \lambda \, {\rm TV}(f), \\
& {\rm TV} \in \{ \TVi, \TVa\} \notag,
\end{align}
with the last expression in (\ref{TVgen}) meaning that the TV regularization can be either isotropic or anisotropic. 

In this paper we are particularly interested in the case when $\Hh$ represents an operator of convolution, in which case the resulting restoration problem becomes that of image deconvolution. More specifically, $\Hh$ is assumed to be {\em mean-preserving} linear filtering, which implies $\Hh\{{\bf 1}\} = {\bf 1}$. Note that (subject to a proper normalization) the class of such {\em blurs} is relatively broad, including the important examples of moving-average, out-of-focus, and motion blurs, just to name a few. The mean-preserving property of $\Hh $ guarantees that $z=0$ is the only vector contained in the intersection of the null spaces of $\|\Hh\{ z \}\|_2: \Rnm \rightarrow \mathbb{R}$ and ${\rm TV}(z): \Rnm \rightarrow \mathbb{R}$, which in turn suggests that, for any $\lambda > 0$, the function $\|\Hh\{z\}\|_2^2 + \lambda \, {\rm TV}(z): \Rnm \rightarrow \mathbb{R}$ is coercive and strictly convex\footnote{It is, in fact, a norm.}. As a result, the optimization problem (\ref{TVgen}) is guaranteed to admit a unique global minimizer~\cite{Aujol09}.

It should also be noted that, in the case of $g \in \UU$ (which can always be enforced by setting the mean value of the data image $g$ to zero), the minimization in (\ref{TVgen}) can be performed over $\Rnm$ (rather than $\UU$), with the solution guaranteed to belong to $\UU$. This fact can be easily verified by contradiction as follows. Assume $z \in \Rnm$ is a global minimizer of $E$ in (\ref{TVgen}). Denoting by $\mu \neq 0$ the mean value of $z$ (i.e. $\mu = (N M)^{-1} \sum_{n,m} z_{n,m}$), the latter can be represented as $z = \tilde{z} + \mu \, {\bf 1}$, where $\tilde{z} \in \UU$. However, due to the orthogonality of $\UU$ w.r.t. the subspace of constant images as well as due to the fact that ${\rm TV}(z)={\rm TV(\tilde{z})}$ , it holds that
\begin{equation*}
E(z) = \frac{1}{2}\left\| \Hh\{ \tilde{z} \} - g \right\|_2^2 + \frac{1}{2}\| \mu \, {\bf 1} \|_2^2 + {\rm TV}(\tilde{z}) = E(\tilde{z}) + N M \mu^2 > E(\tilde{z}), 
\end{equation*}
which contradicts the assumption on $z$ to be a global minimizer.   

The fact that, for $g \in \UU$, the global minimizer
\begin{equation}\label{global}
f_{\rm TV} = \argmin_{f \in \Rnm} \left\{E(f)\right\}
\end{equation}
belongs to $\UU$ will play a key role in the derivations that follow. 
	
\subsection{Gradient and Divergence Operators on $\UU$}
Let $\nabla$ denote the operator of {\em discrete gradient} defined in the standard manner as
\begin{equation}
\nabla: \Rnm \rightarrow \RR: f \mapsto \nabla f = \left(
\begin{array}{c}
f_x \\ 
f_y 
\end{array} 
\right) 
\end{equation}
for some standard definitions of the partial differences $f_x$ and $f_y$ (see below). The method proposed in this paper is based on the fact that {\em the restriction of $\nabla$ to $\UU$ is injective and hence invertible on its image}. Let $\nabla_\UU$ denote the restriction of $\nabla$ to $\UU$ and $\VV := {\rm range}(\nabla_\UU) \equiv {\rm range}(\nabla)$. Then, the above statement suggests that there exists a {\em left inverse} operator
\begin{equation}
\mathcal{U}: \VV \rightarrow \UU:  {\bf v} = \left(
\begin{array}{c}
v_x \\ 
v_y 
\end{array} 
\right) \mapsto u
\end{equation} 
such that 
\begin{equation}\label{leftinv}
\mathcal{U} \left\{ \nabla f \right\} = f,  \,\,\,\, \forall f \in \UU.
\end{equation}
Below we explicitly construct the gradient $\nabla$ and the operator $\mathcal{U}$ for two practically important cases of replicative and periodic boundary conditions. 

\subsubsection{Replicative boundary conditions}\label{rep} In this case the discrete gradient $\nabla f$ can be defined according to
\begin{equation}\label{repGrad}
(\nabla f)_{n,m} =
\begin{cases}
(f_x)_{n,m}  = f_{n,m} - f_{n-1,m}, \mbox{ with } f_{-1,m} = f_{0,m} \\
(f_y)_{n,m}  = f_{n,m} - f_{n,m-1}, \mbox{ with } f_{n,-1} = f_{n,0}
\end{cases}  
\end{equation}
where $n=0,1,\ldots,N-1$ and $m=0,1,\ldots,M-1$. Subsequently, congruent to the definition of (\ref{repGrad}), the {\em discrete divergence} operator ${\rm div}: \RR \rightarrow \Rnm$ can then be defined as
\begin{align}\label{repDiv}
&{\bf v} = \left( \begin{array}{c} v_x \\ v_y \end{array} \right) \mapsto {\rm div}({\bf v}), \mbox{ where } \notag \\ 
\left({\rm div}({\bf v})\right)_{n,m} &= 
\begin{cases}
(v_x)_{1,m}, &\mbox{ if } n=0 \\
(v_x)_{n+1,m} - (v_x)_{n,m}, &\mbox{ if }  0 < n < N-1 \\
- (v_x)_{N-1,m}, &\mbox{ if } n=N-1
\end{cases}
\quad + \\
&+ \, \begin{cases}
(v_y)_{n,1}, &\mbox{ if } m=0 \\
(v_y)_{n,m+1} - (v_y)_{n,m}, &\mbox{ if } 0 < m < M-1 \\
- (v_y)_{n,M-1}, &\mbox{ if } m=M-1
\end{cases} \notag
\end{align}
for any ${\bf v} \in \RR$. 

It is important to point out that the above definitions of the gradient and divergence operators are consistent with their continuous counterparts in the sense that $-{\rm div}$ constitutes the adjoint operator of $\nabla$. Specifically, for any ${\bf v} \in \RR$ and $u \in \Rnm$, it holds that
\begin{equation}\label{adjoint}
\left\langle \nabla u, {\bf v} \right\rangle_{\RR} = \left\langle u, -{\rm div}({\bf v}) \right\rangle,
\end{equation}
where $\left\langle {\bf v}, {\bf w} \right\rangle_{\RR} \equiv \langle v_x, w_x\rangle  + \langle v_y, w_y\rangle$.

Let $\mathcal{DCT}: \mathbb{R}^{N\times M} \rightarrow \mathbb{R}^{N\times M}$ be the operator of 2-D {\em discrete cosine transform} (DCT) (as it can be implemented using, e.g., the {\tt dct2} function of MATLAB). Then, with the definitions (\ref{repGrad}) and (\ref{repDiv}), it is straightforward to show that, for any $u \in \Rnm$
\begin{equation}\label{equality1}
\mathcal{DCT}\{ {\rm div} \left( \nabla u \right) \} = \mathcal{DCT}\{u\} \cdot W,     
\end{equation}
where the dot stands for element-wise matrix product and the elements of the $N\times M$ matrix $W$ are defined as
\begin{equation}\label{repW}
W_{k,l} = 2 \cos\left\{ \frac{\pi k}{N} \right\} + 2 \cos\left\{ \frac{\pi l}{M} \right\} - 4, 
\end{equation}
with $k=0,1,\ldots, N-1$ and $l = 0,1,\ldots,M-1$. It is important to point out that all the values $W_{k,l}$ are strictly positive for $k+l > 0$, while $W_{0,0}=0$. This fact makes it possible to define the {\em integration filter} $W_i \in \Rnm$ as 
\begin{equation}\label{repWi}
(W_i)_{k,l} = 
\begin{cases}
\left[ 2 \cos\left\{ \frac{\pi k}{N} \right\} + 2 \cos\left\{ \frac{\pi l}{M} \right\} - 4 \right]^{-1}, &k+l > 0 \\
0, &k=l=0,
\end{cases}
\end{equation}
which satisfies
\begin{equation}\label{identity}
W_{k,l} \, (W_i)_{k,l} = 1 - \delta(k+l) = 
\begin{cases}
1, \mbox{ if } k+l > 0 \\
0, \mbox{ if } k=l=0.
\end{cases}
\end{equation}
In combination with (\ref{equality1}) and (\ref{identity}), the fact that, for any $u \in \UU$, $\left( \mathcal{DCT}\{u\} \right)_{0,0} = 0$ suggests that
\begin{equation}\label{repGradToFun}
u = \mathcal{DCT}^{-1} \left\{ \mathcal{DCT}\left\{ {\rm div} (\nabla u) \right\} \cdot W_i \right\},
\end{equation}
which, in turn, leads to the definition of operator $\mathcal{U}$ as
\begin{equation}\label{repU}
\mathcal{U}: \RR \rightarrow \UU: {\bf v} \mapsto \mathcal{DCT}^{-1} \left\{ \mathcal{DCT}\{{\rm div}({\bf v})\} \cdot W_i \right\}, 
\end{equation}
Note that $\Uu$ defined by (\ref{repU}) obviously satisfies (\ref{leftinv}).

\subsubsection{Periodic boundary conditions}\label{per} For the sake of completeness, we provide definitions analogous to (\ref{repGrad}), (\ref{repDiv}), and (\ref{repGradToFun}) for the case of periodic boundary conditions. Specifically, in this case, the gradient operator is defined as
\begin{equation}\label{perGrad}
(\nabla f)_{n,m} =
\begin{cases}
(f_x)_{n,m}  = f_{n,m} - f_{n-1,m}, \mbox{ with } f_{-1,m} = f_{N-1,m} \\
(f_y)_{n,m}  = f_{n,m} - f_{n,m-1}, \mbox{ with } f_{n,-1} = f_{n,M-1}
\end{cases}  
\end{equation}
with $n=0,1,\ldots,N-1$ and $m=0,1,\ldots,M-1$, while the corresponding divergence operator is defined as
\begin{align}\label{perDiv}
\left({\rm div}({\bf v})\right)_{n,m} &= 
\begin{cases}
(v_x)_{n+1,m} - (v_x)_{n,m}, \mbox{ if }  0 \leq n < N-1 \\
(v_x)_{0,m} - (v_x)_{N-1,m}, \mbox{ if }  n=N-1
\end{cases}
+ \\
& \, + \begin{cases}
(v_y)_{n,m+1} - (v_y)_{n,m}, \mbox{ if } 0 \leq m < M-1 \\
(v_y)_{n,0} - (v_y)_{n,M-1}, \mbox{ if }  m=M-1
\end{cases}
\end{align}
for any ${\bf v} \in \RR$.

For the above definitions of the gradient and divergence operators it can be shown that, for any $u \in \Rnm$, one has
\begin{equation}\label{equality2}
\mathcal{DFT}\{ {\rm div} \left( \nabla u \right) \} = \mathcal{DFT}\{u\} \cdot W,
\end{equation}
where $\mathcal{DFT}: \Rnm \rightarrow \mathbb{C}^{N\times M}$ is the operator of 2-D {\em discrete Fourier transform} (DFT) (as it can be implemented using, e.g., the {\tt fft2} function of MATLAB) and 
\begin{equation}\label{perW}
W_{k,l} = 2 \cos\left\{ \frac{2 \pi k}{N} \right\} + 2 \cos\left\{ \frac{2 \pi l}{M} \right\} - 4, 
\end{equation}
with $k=0,1,\ldots, N-1$ and $l = 0,1,\ldots,M-1$. 

Similarly to the case of Section~\ref{rep}, the fact that $(\mathcal{DFT}\{u\})_{0,0}=0, \forall u \in \UU$ in conjunction with (\ref{equality2}) leads us to conclude that
\begin{equation}\label{perGradToFun}
u = \mathcal{DFT}^{-1} \left\{ \mathcal{DFT}\left\{ {\rm div} (\nabla u) \right\} \cdot W_i \right\}
\end{equation}
with
\begin{equation}\label{perWi}
(W_i)_{k,l} = 
\begin{cases}
\left[ 2 \cos\left\{ \frac{2 \pi k}{N} \right\} + 2 \cos\left\{ \frac{2 \pi l}{M} \right\} - 4 \right]^{-1}, &k+l > 0 \\
0, &k=l=0.
\end{cases}
\end{equation}
holds for any image $u$ in $\UU$. As a result, the operator $\mathcal{U}$, defined as
\begin{equation}\label{perU}
\mathcal{U}: \RR \rightarrow \UU: {\bf v} \mapsto \mathcal{DFT}^{-1} \left\{ \mathcal{DFT}\{{\rm div}({\bf v})\} \cdot W_i \right\},
\end{equation}
satisfies the condition of (\ref{leftinv}). 

\subsection{Projection on the Range of $\nabla_\VV$}\label{ProjU}
It is important to point out that the definitions of $\mathcal{U}$ in (\ref{repU}) and (\ref{perU}) are nothing else but discrete counterparts of the well-known constructions used for solution of the Poisson equation in continuous variational analysis~\cite{Rektorys80}. Furthermore, the discrete setting makes it straightforward to prove that, for an arbitrary ${\bf v} \in \RR$, the vector $u = \mathcal{U}\{{\bf v}\}$ constitutes a unique minimizer of the norm $\| \nabla u - {\bf v} \|_{\RR}$ among all vectors of $\UU$, {\it viz.}\footnote{Note that the uniqueness of $u=\mathcal{U}\{{\bf v}\}$ as a minimizer follows from the equivalence of the norms $\| \nabla \cdot\|_{\RR}$ and $\| \cdot \|_2$ in $\UU$ due to the special structure of $\UU$ (i.e. ``no constant images are allowed") as well as because of $\UU$ being a finite dimensional subspace.} 
\begin{align}\label{min}
\mathcal{U}\{{\bf v}\} &= \argmin_{u \in \UU} \left\{ \left\| \nabla u - {\bf v} \right\|_{\RR}^2 \right\}  = \notag \\
&=\argmin_{u \in \UU} \left\{ \left\| u_x - v_x \right\|_2^2 + \left\| u_y - v_y \right\|_2^2 \right\}, \,\, \forall {\bf v} \in \RR
\end{align}
Consequently, (\ref{min}) suggests that the composite operator $\nabla \mathcal{U}$ defined as 
\begin{equation}\label{proj}
\nabla \mathcal{U} : \RR \rightarrow \VV: {\bf v} \mapsto \nabla \left\{ \mathcal{U}\{{\bf v}\} \right\}
\end{equation}
constitutes an orthogonal projection of $\RR$ onto  $\VV$. (The property (\ref{proj}) of $\mathcal{U}$, in fact, defines it as a left inverse.) It is also interesting to note that, considering ${\bf v} \in \RR$ to be a vector field, the operator $\nabla \mathcal{U}\{{\bf v}\}$ sets to zero the rotational (solenoidal) component of $\bf v$, and therefore $\nabla \mathcal{U}$ is, in fact, an orthogonal projector onto the subspace of irrotational (curl-free) vector fields~\cite{Arfken95}.

\section{Derivative Shrinkage}
Similarly to the case with many other image restoration problems, the solution of (\ref{TVgen}) can be given a statistical interpretation~\cite{Li09}. Particularly, from the viewpoint of Bayesian estimation, the isotropic regularizer $\TVi$ (\ref{iTV}) favors solutions $f$ with independent and identically distributed values $(\nabla f)_{n,m}$. Moreover, at each pixel $(n,m)$, the phase $\arctan(f_y/f_x)_{n,m}$ of $(\nabla f)_{n,m}$ is assumed to be uniformly distributed in $(-\pi, \pi]$, while its magnitude $| (\nabla f)_{n,m} |$ follows a Laplacian distribution. The anisotropic regularizer $\TVa$ (\ref{aTV}), on the other hand, suggests that the partial differences $f_x$ and $f_x$ are mutually independent and identically distributed according to a Laplacian law. Needless to say that the reconstructions corresponding to $\TVi$ and $\TVa$ generally differ in their properties and appearance. In this paper, a unified solution will be given to address both above cases. For methodological reasons, however, the case of the anisotropic TV (\ref{aTV}) will be worked out first.

Let ${\bf f} = (f_x, \ f_y )^T \in \VV$ denote the gradient of $f$, and hence, with a slight abuse of nomenclature, one can say $\mathcal{U}\{\nabla f\} \equiv \mathcal{U}\{ {\bf f} \}$. With the use of this new notation, we first notice that
\begin{equation}
\|{\bf f}\|_1 = \sum_{n,m} |(f_x)_{n,m}| + |(f_y)_{n,m}| = \TVa(f),
\end{equation}
and, hence, the optimization problem (\ref{TVgen}) (with ${\rm TV}=\TVa$) can be replaced by an equivalent problem of the form
\begin{equation}\label{aTVd}
{\bf f}_{\TVa} = \argmin_{{\bf f} \in \VV} \left\{  \frac{1}{2} \| \Hh\left\{ \mathcal{U}\{ {\bf f} \} \right\} - g \|_2^2 + \lambda \| {\bf f} \|_1 \right\}
\end{equation}
with the operator $\mathcal{U}$ defined by either (\ref{repU}) or (\ref{perU}). In other words, in (\ref{aTVd}) the minimization over $f\in \UU$ is replaced by minimization over its partial differences ${\bf f} \in \VV$. Note that the equivalence between the original problem and (\ref{aTVd}) is underpinned by the existence of $\mathcal{U}$ that defines a one-to-one correspondence between $\VV$ and the signal space $\UU$. It should be pointed out that, as long as the minimization domain in (\ref{aTVd}) is restricted to be $\VV$, the cost functional in (\ref{aTVd}) remains coercive and strictly convex, which in turn guarantees the existence of a unique global minimizer in $\UU$. Moreover, ${\bf f}_{\TVa}$ can be used to recover the solution $f_{\TVa}$ to the original problem (\ref{TVgen}) according to
\begin{equation}
f_{\TVa} = \mathcal{U} \{{\bf f}_{\TVa}\}.
\end{equation}

The form of (\ref{aTVd}) can be additionally simplified via introducing the operator $\Aa: \VV \rightarrow \UU$ as a composition of the convolution $\Hh$ and integration $\mathcal{U}$ operators, i.e. $\Aa \{\cdot\} := \Hh\{ \mathcal{U} \{ \cdot \}\}$. Using $\Aa$ allows (\ref{aTVd}) to be rewritten in a more standardized form as
\begin{align}\label{L1cost}
{\bf f}_{\TVa} = \argmin_{{\bf f} \in \VV} \left\{ E_{\TVa} ({\bf f}) \right\}, \mbox{ where }\\
E_{\TVa} ({\bf f}) =  \frac{1}{2} \| \Aa\{ {\bf f} \} - g \|_2^2 + \lambda \| {\bf f} \|_1. \notag
\end{align}

The problem (\ref{L1cost}) has a format identical to that of the sparse-constrainted reconstruction problems of~\cite{Figueiredo03, Daubechies03}, and hence it can be solved by the method of iterative shrinkage. The fact, however, that $E_{\TVa}$ in (\ref{L1cost}) is minimized over $\VV$ (rather than over $\RR$) makes it necessary to supplement each step of the iterative shrinkage by the orthogonal projection onto $\VV$ according to (\ref{proj}). The resulting algorithm can be summarized as follows. Let $\mathcal{S}_\tau$ (with $\tau > 0$) be the operator of thresholding defined in the standard way as
\begin{equation}\label{Soft}
S_\tau(x) = {\rm sign}(x) \left( |x| - \tau \right)_+.
\end{equation}
Then the proposed algorithm for iteratively solving (\ref{L1cost}) finds ${\bf f}_{\TVa}$ as a stationary point of the sequence of estimates produced by the following iterations
\begin{align}\label{ISa}
{\bf f}^{\left(t+\frac{1}{2}\right)} &= \mathcal{S}_{\lambda/c} \left\{ {\bf f}^{(t)} + \frac{1}{c} \Aa^\ast \left\{ g - \Aa\{{\bf f}^{(t)}\}\right\} \right\} \\
{\bf f}^{(t+1)} &= \nabla \mathcal{U} \{ {\bf f}^{\left(t+\frac{1}{2}\right)} \},  \notag
\end{align}
where $\Aa^\ast$ is the adjoint operator of $\Aa$, and $c$ is a positive scalar obeying  $c > \| \Aa \Aa^\ast \|$. The structure of (\ref{ISa}) can be additionally simplified by introducing ${\bf b} := \Aa^\ast \{g\}$ and $\Rr(\cdot) := \Aa^\ast \{\Aa\{\cdot\}\}$. In this case, the iteration procedure (\ref{ISa}) can be rewritten more concisely as
\begin{equation}\label{Alga}
{\bf f}^{(t+1)} = \nabla \Uu \left\{ \mathcal{S}_{\lambda/c} \left\{ {\bf f}^{(t)} + c^{-1} \left( {\bf b} - \Rr\{{\bf f}^{(t)}\} \right) \right\} \right\},
\end{equation}
Note that (\ref{Alga}), in fact, defines a map from $\VV$ to itself, which can be shown to be non-expansive and asymptotically regular (see Remark 3.12 in~\cite{Daubechies03}). In combination with the property of $\VV$ being a convex and closed set, the above fact guarantees that the iterations in (\ref{Alga}) converge to a minimizer of $E_{\TVa}$~\cite[Prop. 3.9]{Daubechies03}.

To complete the description of the iterative shrinkage algorithm for $\TVa$-based image restoration, the operators $\Aa^\ast$ and $\Rr$ should be explicitly defined along with the constant $c$. To this end, we first recall that $\Aa$ is defined as a composition of the operators $\Hh$ and $\Uu$, {\it viz.}
\begin{equation}\label{Adef}
\Aa: \VV \rightarrow \UU: {\bf v} \mapsto  \Hh\left\{ \fft^{-1}\left\{ \fft \left\{{\rm div}({\bf v})\right\} \cdot W_i \right\} \right\}
\end{equation}
The above operation can be implemented at the cost of an FFT-based convolution if the operator $\Hh$ corresponds to {\em periodic} convolution. In this case, $\Hh$ can be represented by an $N\times M$ matrix $H$ of the DFT of its associated convolution kernel, which allows the operator $\Aa$ to be defined as
\begin{equation}\label{Aper}
\Aa: \VV \rightarrow \UU: {\bf v} \mapsto  \fft^{-1}\left\{ \fft \left\{{\rm div}({\bf v})\right\} \cdot A \right\},
\end{equation}
with $A$ being the frequency response of the composition of $\Hh$ and integration $W_i$, i.e.
\begin{equation}\label{AWiH}
A = W_i \cdot H.
\end{equation}
It should be noted that periodic boundary conditions are common in image processing, since they allow substantially reducing the computational cost of filtering-type operations through the use of FFT. Moreover, there is a number of standard techniques which make it possible to adapt arbitrary images for the processing by periodic convolution~\cite{Nagy96}. For these reasons, the derivations below will be confined to the case of periodic convolution with the operator $\Aa$ defined by (\ref{Aper}) and (\ref{AWiH}).

To specify the adjoint operator $\Aa^\ast$ of $\Aa$, let $\bf v$ and $u$ be two arbitrary vectors in $\VV$ and $\UU$, respectively. Then, 
\begin{equation}
\left\langle \Aa\{{\bf v}\}, u\right\rangle = \left\langle {\bf v}, \Aa^\ast\{u\}\right\rangle= \left\langle {\rm div}({\bf v}), \fft^{-1} \left\{ \fft\{u\} \cdot \bar{A} \right\}\right\rangle,
\end{equation}
and, therefore,
\begin{equation}\label{Astar}
\Aa^\ast: \UU \rightarrow \VV: u \mapsto -\nabla \left( \fft^{-1} \left\{ \fft\{u\} \cdot \bar{A} \right\} \right),
\end{equation}
where $\bar{A}$ denotes the complex conjugate of $A$. Note that, using the above definition of $\Aa^\ast$, the vector $\bf b$ in (\ref{Alga}) can be (pre-)computed according to
\begin{equation}\label{BB}
{\bf b} = \Aa^\ast\{g\} = - \nabla \left( \fft^{-1} \left\{ \fft\{g\} \cdot \bar{A} \right\} \right),
\end{equation}
in which case it appears to be unnecessary to preprocess $g$ by setting to zero its mean value, as the ``DC" component of $\fft\{g\}$ is, in any event, multiplied by $(\bar{A})_{0,0}=0$. Moreover, by direct substitution one obtains
\begin{equation}\label{Rper}
\Rr:  \VV \rightarrow \VV: {\bf v} \mapsto  -\nabla \left(  \fft^{-1} \left\{  \fft \left\{{\rm div}({\bf v}) \cdot |A|^2 \right\} \right\} \right),
\end{equation}
with $|A|^2 = A \cdot \bar{A}$. Note that the computational cost of applying $\Rr$ is actually defined by the cost of one FFT-based convolution.

Finally, to determine the range of admissible values of the parameter $c$ in (\ref{Alga}), we first note that, for an arbitrary $u \in \UU$, it holds that
\begin{equation}\label{AAA}
\Aa\{\Aa^\ast\{u\}\} = \fft^{-1} \left\{ \fft\{u\} \cdot |A|^2 \cdot W \right\}
\end{equation}
with $W$ given by (\ref{perW}). This suggests that the composition $\Aa \Aa^\ast$ corresponds to convolution of an input image with $\fft^{-1}\{|A|^2\cdot W\}$. Consequently, using the fact that $|A|^2 \cdot W = W_i \cdot |H|^2$, one has
\begin{equation}
\| \Aa \Aa^\ast \| = \max_{n,m} \, (|W_i| \cdot |H|^2)_{n,m} \leq \max_{n,m} \, (|W_i|)_{n,m} \max_{n,m} (|H|^2)_{n,m},
\end{equation}
Moreover, using the definition (\ref{perWi}), it is straightforward to show that
\begin{equation}\label{maxWi}
\max_{n,m} \, (|W_i|)_{n,m} = \left[ 2 - 2 \cos \left( \frac{2\pi}{\max\{N,M\}} \right) \right]^{-1}.
\end{equation}
Consequently, subject to the normalization $\max_{n,m} (|H|^2)_{n,m} = 1$, it follows that the admissible values of $c$ should obey $c > \left[ 2 - 2 \cos \left( 2\pi \slash \max\{N,M\} \right) \right]^{-1}$.

Algorithm 1 below provides an outline of the proposed method for $\TVa$-based image restoration through iterative shrinkage. (Note that it is assumed that the blur operator has been normalized to have $\max_{n,m} (|H|^2)_{n,m} = 1$) The primary purpose of Algorithm 1 is to connect together the most important results on this section, while more general versions of the method will be discussed in the sections that follow.

\begin{algorithm}
\caption{$\TVa$-based image restoration by iterative shrinkage} 
\begin{algorithmic}[1]
\STATE $c \Leftarrow \left[ 2 - 2 \cos \left( 2\pi \slash \max\{N,M\} \right) \right]^{-1} + \epsilon$ (for some $\epsilon > 0$) 
\STATE ${\bf b} \Leftarrow \Aa^\ast\{g\}$ (using (\ref{BB}))
\STATE ${\bf f} \Leftarrow \nabla g$ (using (\ref{perGrad}))
\WHILE{``$\bf f$ keeps changing"}
\STATE ${\bf f} \Leftarrow \mathcal{S}_{\lambda/c} \left\{ {\bf f} + c^{-1} \left( {\bf b}-\Rr\{{\bf f}\} \right) \right\}$ (using (\ref{Soft}) and (\ref{Rper}))
\STATE ${\bf f} \Leftarrow \nabla \Uu\{{\bf f}\}$ (using (\ref{perGrad}) and (\ref{perU}))
\ENDWHILE
\STATE $f_{\TVa} \Leftarrow \Uu\{{\bf f}\}$ (using (\ref{perU}))
\STATE Re-normalize $f_{\TVa}$ (optional)
\end{algorithmic}
\end{algorithm}

\section{Extension to the Case of Isotropic TV}
\subsection{Multidirectional Gradient}
The method of the previous section has been derived for the TV-regularizer in (\ref{TVgen}) equal to $\TVa$ as given by (\ref{aTV}). It is known, however, that using the isotropic TV-regularizer (\ref{iTV}) can provide qualitatively different solutions to the restoration problem. It is, therefore, tempting to extend the results of the preceding sections to the case of isotropic TV regularization, i.e. ${\rm TV} = \TVi$.

To find a connection between $\TVa$ and $\TVi$, we take advantage of the following identity
\begin{equation}\label{integral}
\frac{1}{2} \int_0^{\pi/2} \left( \left| a \cos\theta + b \sin\theta \right| + \left| b \cos\theta - a \sin\theta \right| \right) d\theta = \sqrt{a^2 + b^2},
\end{equation}
which holds for any pair of real numbers $a$ and $b$. Using the fact that $\int_0^{\pi/2} \left( \cos\theta + \sin\theta \right) d\theta =2$, the equality (\ref{integral}) can be alternatively expressed as
\begin{equation}\label{int2}
\frac{\int_0^{\pi/2} \left( \left| a \cos\theta + b \sin\theta \right| + \left| b \cos\theta - a \sin\theta \right| \right) d\theta}{\int_0^{\pi/2} \left( \cos\theta + \sin\theta \right) d\theta} = \sqrt{a^2 + b^2}.
\end{equation}

Now, let $\bar{\theta}_L=\{\theta_0^L, \theta_1^L, \ldots, \theta_{L-1}^L\}$ be a set of $L$ points uniformly distributed in $[0, \pi/2)$. In particular, we define $\theta_k^L = \pi k/2 L$, with $k=0,1,\ldots,L-1$. These points can be used to construct a Riemannian approximation $I(a,b; L)$ to (\ref{integral}) (or, equivalently, to (\ref{int2})) as given by  
\begin{align}
I(a,b; L) =  \frac{\sum_{k=0}^{L-1}\left(\left|a \cos\theta_k^L + b \sin\theta_k^L\right| + \left|b \cos\theta_k^L - a \sin\theta_k^L\right|\right) (\pi/2L)}{\sum_{k=0}^{L-1}(\cos\theta_k^L + \sin\theta_k^L) (\pi/2L)} = \\
=  \frac{\sum_{k=0}^{L-1}\left(\left|a \cos\theta_k^L + b \sin\theta_k^L\right| + \left|b \cos\theta_k^L - a \sin\theta_k^L\right|\right)}{\sum_{k=0}^{L-1}(\cos\theta_k^L + \sin\theta_k^L)}. \notag
\end{align}
Due to the continuity of the integrands in (\ref{int2}), the Riemannian approximation is guaranteed to converge to $\sqrt{a^2+b^2}$ as $L$ goes to infinity. Formally, 
\begin{equation}\label{limit}
\lim_{L \rightarrow \infty} I(a,b; L) = \sqrt{a^2 + b^2}.
\end{equation}
Moreover, when considered as a function of $(a,b) \in \mathbb{R}^2$, $I(a,b; L)$ represents an upper bound on $\sqrt{a^2 + b^2}$ for any value $L \geq 1$. This fact can be formalized in the following lemma.

\newtheorem{lem}{Lemma}
\begin{lem}
For any $L \geq 1$ and any $(a,b) \in \mathbb{R}^2$: 
\begin{equation}
I(a,b; L) \geq  \sqrt{a^2 + b^2},
\end{equation}
while the equality holds only for those $(a,b)$ which satisfy either
\begin{equation}
a \cos\theta_k^L + b \sin\theta_k^L = 0 \mbox{ or } -a \sin\theta_k^L + b \cos\theta_k^L = 0,
\end{equation}
where $\theta_k^L = \pi k/2 L$, with $k=0,1,\ldots,L-1$.
\end{lem}

\begin{figure}[top]
\centering
\includegraphics[width=5in]{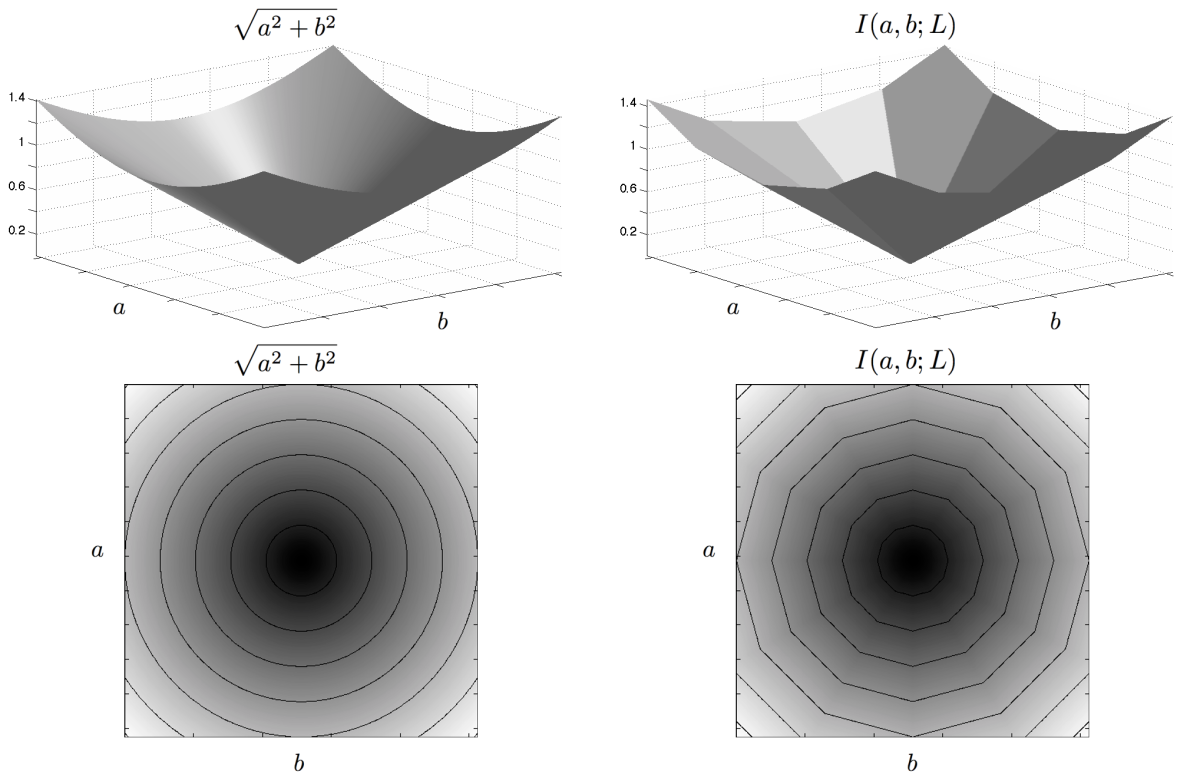}
\caption{The upper row of subplots compare the functions $\sqrt{a^2+b^2}$ and $I(a,b; L=3)$ via visualizing them as surface plots. The lower row of subplots show the same functions as gray-scale images superimposed by their corresponding level set contours.}
\label{F1}
\end{figure}

The proof of Lemma 1 is straightforward, yet technical. For the reason of space, therefore, the proof is omitted here. Instead, Fig.~\ref{F1} compares the functions $\sqrt{a^2+b^2}$ and $I(a,b;L)$ (for the case $L=3$) visualizing them as both surface plots (upper rows of subplots in Fig.~\ref{F1}) and gray-scale images superimposed by their corresponding level set contours (lower rows of subplots in the same figure.) One can see that, as suggested by the lemma, the functions $\sqrt{a^2 + b^2}$ and $I(a,b; L)$ coincide along the directions in $\mathbb{R}^2$ which are defined by the angles $\{\theta_k^L\}_{k=0}^{L-1}$ and $\{\theta_k^L+\pi/2\}_{k=0}^{L-1}$. It is also interesting to note that the level sets of $I(a,b; L)$ provide a piecewise linear approximation to the level sets of $\sqrt{a^2+b^2}$, which becomes progressively more accurate as $L$ increases. 

Using the approximating function $I(a,b;L)$ allows an alternative definition of $\TVi$ as follows
\begin{equation}
\TVi(f) = \left\langle {\bf 1}, \sqrt{f_x^2 + f_y^2} \right\rangle = \lim_{L\rightarrow \infty} \Big\langle {\bf 1}, I(f_x,f_y; L) \Big\rangle, 
\end{equation}
which, in turn, leads to the following approximation of $\TVi$
\begin{align}\label{TVapprox}
&\TVi(f) \simeq \Big\langle {\bf 1}, I(f_x,f_y; L) \Big\rangle = \\
&= d_L \, \Big\langle {\bf 1}, \sum_{k=0}^{L-1} \left|f_x \cos\theta_k^L + f_y \sin\theta_k^L \right| + \left|f_y \cos\theta_k^L - f_x \sin\theta_k^L \right| \Big\rangle \notag
\end{align}
with
\begin{equation}\label{factor}
d_L = \left[ \sum_{k=0}^{L-1} \left( \cos\theta_k^L + \sin\theta_k^L \right) \right]^{-1}.
\end{equation}
The quality of the approximation (\ref{TVapprox}) is supposed to become progressively better as $L$ increases. In the experimental part of the paper, however, it will be shown experimentally that setting $L=3$ results in image restoration practically indistinguishable from the case of exact $\TVi$. It is also important to note that, for the case $L=1$, one has
\begin{equation}
\Big\langle {\bf 1}, I(f_x,f_y; L) \Big\rangle \Big|_{L=1} = \Big\langle {\bf 1} , |f_x| + |f_y| \Big\rangle  = \TVa(f).
\end{equation}
In other words, setting $L=1$ transforms the approximative TV functional into (\ref{TVapprox}) into $\TVa$. For the convenience of future referencing, let the approximative TV functional be denoted by $\TVL$, {\it viz.}
\begin{align}\label{TVL}
&\TVL(f) = \Big\langle {\bf 1}, I(f_x,f_y; L) \Big\rangle = \\
&= d_L \, \Big\langle {\bf 1}, \sum_{k=0}^{L-1} \left|f_x \cos\theta_k^L + f_y \sin\theta_k^L \right| + \left|f_y \cos\theta_k^L - f_x \sin\theta_k^L \right| \Big\rangle \notag
\end{align}
Then, the principal properties of $\TVL$ can be summarized as
\begin{align}\label{bound1}
\TVi(f) &\leq \TVL(f) \leq \TVa(f) \\ 
\forall L &\geq 1, \,\, \forall f\in \Rnm, \notag
\end{align}
and 
\begin{equation}\label{bound2}
\TVL(f) = 
\begin{cases}
\TVa(f), \mbox{ if } L=1 \\
\TVi(f), \mbox{ if } L \rightarrow \infty
\end{cases} 
\end{equation}
Note that (\ref{bound1}) follows as a direct consequence of Lemma 1, while (\ref{bound2}) is due to (\ref{limit}).

Finalizing this subsection, it is instructive to take a closer look at the structure of the summands in the definition (\ref{TVL}) of $\TVL$, which have the form of $\left|f_x \cos\theta_k^L + f_y \sin\theta_k^L \right| + \left|f_y \cos\theta_k^L - f_x \sin\theta_k^L \right|$. Let $\nabla f$, as previously, be the discrete gradient of some $f \in \Rnm$ computed according to either (\ref{repGrad}) or (\ref{perGrad}). Now, let us {\em rotate} $\nabla f$ through the angle $\theta_k^L$ in the counter-clockwise direction. Such a rotated gradient $\nabla_{\theta_k^L} f$ can be expressed via $\nabla f$ and $\theta_k^L$, and it is given by
\begin{equation}\label{rotGrad}
\nabla_{\theta_k^L} f = 
\left(
\begin{array}{c}
f_x \cos\theta_k^L + f_y \sin\theta_k^L \\
f_y \cos\theta_k^L - f_x \sin\theta_k^L
\end{array}
\right).
\end{equation}
Subsequently, given a set of $L$ different angles $\bar{\theta}_L = \{\theta_k^L\}_{k=0}^{L-1}$, one can compute the rotated gradients corresponding to each of the given $\theta_k^L$. Formally, we introduce the notion of {\em multidirectional gradient} (MDG) which is defined as the map $\GRD$ given by 
\begin{equation}\label{GRD}
\GRD: \Rnm \rightarrow \RRL: f \mapsto 
\left(
\begin{array}{c}
\nabla_{\theta_0^L} f \\
\nabla_{\theta_1^L} f \\
\ldots \\
\nabla_{\theta_{L-1}^L} f
\end{array}
\right)
\end{equation}
Now, let ${\bf f}$ be the MDG of $f$, i.e. ${\bf f} = \GRD f = \left( \nabla_{\theta_0^L} f, \nabla_{\theta_1^L} f, \ldots, \nabla_{\theta_{L-1}^L} f \right)^T$. Then, the $\ell_1$-norm of ${\bf f}$ in $\RRL$ can be defined in the standard manner as
\begin{equation}
\|{\bf f}\|_1 = \sum_{n,m} \sum_{k=0}^{L-1} \left(\left|f_x \cos\theta_k^L + f_y \sin\theta_k^L \right| + \left|f_y \cos\theta_k^L - f_x \sin\theta_k^L \right| \right)_{n,m},
\end{equation}
in which case (\ref{TVL}) suggests that 
\begin{equation}\label{TVL1}
\TVL(f) = d_L \| {\bf f} \|_1.
\end{equation}

The relation (\ref{TVL1}) establishes a connection between the $\TVL$ functional and the $\ell_1$-norm which is necessary to derive a method for $\TVL$-based image restoration by means of iterative shrinkage, which is detailed below.

\subsection{$\TVL$-based Reconstruction via Iterative Shrinkage}
The fact that the MDG $\GRD$ depends linearly on $\nabla$ suggests that the null space of $\GRD$ consists of the subset of all constant images in $\Rnm$, and, therefore, both $\GRD$ and its restriction to $\UU$ share the same range which we denote by $\VV_L$, i.e. $\VV_L := {\rm range}(\GRD)$. Note that by the definition of the range, for any ${\bf v} \in \VV_L$ there exists $u \in \UU$ such that $\GRD u = {\bf v}$. Running a few steps forward, let us assume that the restriction of $\GRD$ to $\UU$ is injective, and hence invertible on its image. In such a case, $\GRD$ has to have a left inverse $\UL: \VV_L \rightarrow \UU$ whose defining properties are
\begin{enumerate}
\item $\UL\{ \GRD f \} = f$, for all $f \in \UU$,
\item For any ${\bf v} \in \RRL$, $\argmin_{u \in \UU} \left\| \GRD u - {\bf v} \right\|_{\RRL}^2 = \UL\{{\bf v}\}$.  
\end{enumerate}
Consequently, using the operator $\UL$ one can replace the original restoration problem (\ref{TVgen}) with ${\rm TV} = \TVL$ by an equivalent problem
\begin{equation}\label{grandTV}
{\bf f}_{\TVL} = \argmin_{{\bf f} \in \VV_L} \left\{ \frac{1}{2} \left\| \Hh\left\{ \mathcal{U}_L\{{\bf f}\} \right\} - g \right\|_2^2 + (\lambda \, d_L)  \| {\bf f} \|_1 \right\},
\end{equation}
which appears to be in the format suitable for its solution via iterative shrinkage~\cite{Figueiredo03, Daubechies03}. Note that given ${\bf f}_{\TVL}$, the corresponding estimate $f_{\TVL}$ of $f$ can be computed according to
\begin{equation}
f_{\TVL} = \mathcal{U}_L\{ {\bf f}_{\TVL} \},
\end{equation}
which coincides with $f_{\TVa}$ for $L=1$ and approaches $f_{\TVi}$ when $L$ increases. Therefore, by merely varying the value of $L$, (\ref{grandTV}) allows switching between the anisotropic and isotropic TV reconstructions.

Throughout the rest of this section, we construct the operator $\UL$ and provide the definitions of operators analogous to the operators $\Aa$, $\Aa^\ast$, and $\Rr$ of the preceding section. To proceed with the derivations, it is first necessary to find the divergence operator $\DIV$ which is congruent with the definition of $\GRD$. Such operator $\DIV$ can be computed based on its property as an adjoint operator which requires that
\begin{equation}
\langle \GRD u, {\bf v} \rangle_{\RRL} = \langle u, -\DIV {\bf v} \rangle
\end{equation}
holds for any $u \in \Rnm$ and ${\bf v} \in \RRL$. Specifically, straightforward computations lead to the definition of $\DIV$ as
\begin{equation}\label{DIV}
\DIV: \RRL \rightarrow \Rnm: {\bf v} \mapsto -{\rm div}
\left(
\begin{array}{c}
\sum_{k=0}^{L-1} (v_x^k \cos\theta_k^L - v_y^k \sin\theta_k^L) \\
\sum_{k=0}^{L-1} (v_y^k \cos\theta_k^L + v_x^k \sin\theta_k^L)
\end{array}
\right)
\end{equation}
where ${\rm div}$ is given by either (\ref{repDiv}) or (\ref{perDiv}) in compliance with the corresponding definition of $\nabla$, and the elements of ${\bf v}$ are assumed to be ordered as ${\bf v} = \left(v_x^0, v_y^0, v_x^1, v_y^1, \ldots, v_x^{L-1}, v_y^{L-1} \right)$.

Given $\GRD$ and $\DIV$, the operator $\UL$ can be found based on its property as a projection operator. Specifically, let ${\bf v}$ be an arbitrary element in $\RRL$. Then, the vector $u\in \UU$ that minimizes the norm $\left\| \GRD u - {\bf v} \right\|_{\RRL}^2$ should solve the systems of corresponding {\em normal equations}, which can be defined in the operator form as~\cite{Gohberg81}
\begin{equation}\label{normeq1}
\DIV \left( \GRD u \right) = \DIV({\bf v}). 
\end{equation}
To solve the above system we will need the result of the following lemma.

\begin{lem}
Let $L \geq 1$, and let $\GRD$ and $\DIV$ be the operators defined by (\ref{GRD}) and (\ref{DIV}) with respect to $\nabla$ and ${\rm div}$ given by either (\ref{repGrad}) and (\ref{repDiv}) or (\ref{perGrad}) and (\ref{perDiv}), respectively. Then, for any $u \in \UU$
\begin{equation}\label{LEM}
\DIV \left( \GRD u \right) = L \, {\rm div}(\nabla u).
\end{equation}
\end{lem}

Lemma 2 is proven via direct substitution of the definitions of $\GRD$ and $\DIV$ in the left-hand side of (\ref{LEM}) with the use of some standard trigonometric equalities. Again, for the reason of space, the proof is omitted here.

Using the result of Lemma 2 allows the normal equations (\ref{normeq1}) to be expressed in a different, yet equivalent form, {\it viz.}
\begin{equation}\label{normeq2}
{\rm div}(\nabla u) = L^{-1} \DIV({\bf v}).
\end{equation}
Subsequently, depending on the type of boundary conditions, a $u$ satisfying (\ref{normeq2}) can be found by means of either DCT or DFT transforms by virtue of the relations in (\ref{equality1}) or (\ref{equality2}). For the sake of concreteness, let the boundary conditions to be of the periodic type. In this case, the vector $u \in \UU$ uniquely minimizing $\left\| \GRD u - {\bf v} \right\|_{\RRL}^2$ is found to be
\begin{equation}\label{grandsol}
u = \frac{1}{L} \fft^{-1} \left\{ \fft\{ \DIV({\bf v}) \} \cdot W_i \right\},
\end{equation}
where $W_i$ is given by (\ref{perWi}). It should be emphasized that the uniqueness of the above solution is guaranteed by the fact that $u$ is restricted to be an element of $\UU$.  

Based on (\ref{grandsol}), the left inverse operator $\UL$ can now be defined as
\begin{equation}\label{grandU}
\UL: \RRL \rightarrow \UU: {\bf v} \mapsto \frac{1}{L} \fft^{-1} \left\{ \fft\{ \DIV({\bf v}) \} \cdot W_i \right\}. 
\end{equation}
Moreover, following the same line of considerations as in Section~\ref{ProjU}, the operator of orthogonal projection from $\RRL$ onto $\VV_L = {\rm range}(\GRD)$ has the form of
\begin{equation}\label{grandProj}   
\GRD \UL: \RRL \rightarrow \VV_L: {\bf v} \mapsto \GRD \UL\{{\bf v}\}.
\end{equation}

As was already argued before, a considerable gain in computational efficiency becomes possible if the convolution operator $\Hh$ is defined to be periodic. In this case, one can define $\Aa_L$ to be the composition operator given by
\begin{equation}\label{grandA}
\Aa_L: \VV_L \rightarrow \UU: {\bf v} \mapsto \Hh \left\{ \UL\{{\bf v}\} \right\} = \frac{1}{L} \fft^{-1} \left\{ \fft\{ \DIV({\bf v}) \} \cdot A \right\},
\end{equation}
with $A$ defined by (\ref{AWiH}). This allows the optimization problem (\ref{grandTV}) to be redefined in a more standardized way as
\begin{equation}\label{GTV}
{\bf f}_{\TVL} = \argmin_{{\bf f} \in \VV_L} \left\{ \frac{1}{2} \left\| \Aa_L\{{\bf f}\} - g \right\|_2^2 + (\lambda \, d_L) \, \| {\bf f} \|_1 \right\}.
\end{equation}

To find the solution ${\bf f}_{\TVL}$ of (\ref{GTV}), the adjoint operator $\Aa_L^\ast$ of $\Aa_L$ and the composition $\Rr_L(\cdot) := \Aa_L^\ast\{\Aa_L\{ \cdot \} \}$ need to be specified next. These operators can be shown to be respectively given by
\begin{equation}\label{grandAstar}
\Aa_L^\ast: \UU \rightarrow \VV_L:  u \mapsto -\frac{1}{L} \GRD \left(  \fft^{-1} \left\{ \fft\{u\} \cdot \bar{A}\right\}  \right) 
\end{equation}
and 
\begin{equation}\label{grandR}
\Rr_L: \VV_L \rightarrow \VV_L:  {\bf v} \mapsto -\frac{1}{L^2} \GRD \left(  \fft^{-1} \left\{ \fft\{ \DIV({\bf v}) \} \cdot |A|^2 \right\}  \right), 
\end{equation}
where, as previously, $\bar{A}$ stands for the complex conjugate of $A$ and $|A|^2 = A \cdot \bar{A}$. It is important to note that, despite the use of $L$ different gradient directions, the computation cost of applying $\Rr_L$ is still dominated by the cost of only {\em one} FFT-based convolution.

Subsequently, given the above definitions and the theoretical guarantees of~\cite{Daubechies03}, the solution to (\ref{GTV}) can be found iteratively by the following recursion
\begin{equation}\label{grandAlg}
{\bf f}^{(t+1)} = \GRD \UL \left\{ \mathcal{S}_{\frac{\lambda d_L}{c}} \left\{ {\bf f}^{(t)} + c^{-1} \left( {\bf b}_L - \Rr_L \{{\bf f}^{(t)} \} \right) \right\} \right\}
\end{equation}
where the soft thresholding $\mathcal{S}_\tau$ is defined by (\ref{Soft}), ${\bf b}_L := \Aa_L^\ast\{g\}$, and $c$ is required to satisfy $c > \| \Aa_L \Aa_L^\ast \| $. It should be noted that the latter condition is crucial for the convergence of the iterative shrinkage procedure of (\ref{grandAlg}). To determine a range of admissible values of $c$, we first notice that, for any arbitrary $u \in \Rnm$, one has
\begin{equation}\label{Onorm}
\Aa_L \left\{ \Aa_L^\ast \{ u \} \right\} = -\frac{1}{L} \fft^{-1} \left\{ \fft\{u\} \cdot W \cdot |A|^2 \right\},
\end{equation}  
which, by comparison with (\ref{AAA}), leads us to conclude that
\begin{equation}\label{Norms}
\| \Aa_L \Aa_L^\ast \| = \frac{1}{L}\| \Aa \Aa^\ast \|.
\end{equation}
Therefore, provided the convolution blur is normalized to obey $\max_{n,m}(|H|^2)_{n,m}=1$, and based on (\ref{maxWi}), one can conclude that $c$ should be chosen to satisfy
\begin{equation}\label{Lbound}
c > \frac{1}{L} \left[ 2 - 2 \cos \left( \frac{2\pi}{\max\{N,M\}} \right) \right]^{-1}.
\end{equation}

The central results of this section are summarized in Algorithm 2 below. Although the main purpose of Algorithm 2 it to establish connections between the main theoretical results of the paper, it can also be regarded as a ``working prototype" of the proposed method.

\begin{algorithm}
\caption{$\TVL$-based image restoration by iterative shrinkage} 
\begin{algorithmic}[1]
\STATE $\bar{\theta}_L \Leftarrow \{ \pi k \slash 2 L \}_{k=0}^{L-1}$
\STATE $d_L \Leftarrow \left[ \sum_{k=0}^{L-1} \left( \cos\theta_k^L + \sin\theta_k^L \right) \right]^{-1}$
\STATE $c \Leftarrow (1/L) \left[ 2 - 2 \cos \left( 2\pi \slash \max\{N,M\} \right) \right]^{-1} + \epsilon$ (for some $\epsilon > 0$)
\STATE $\tau \Leftarrow (\lambda \, d_L) \slash c$
\STATE ${\bf b}_L \Leftarrow \Aa_L^\ast\{g\}$ (using (\ref{grandAstar}))
\STATE ${\bf f} \Leftarrow \GRD g$ (using (\ref{GRD}))
\WHILE{``$\bf f$ keeps changing"}
\STATE ${\bf f} \Leftarrow \mathcal{S}_\tau \left\{ {\bf f} + c^{-1} \left( {\bf b}_L - \Rr_L\{{\bf f}\} \right) \right\}$ (using (\ref{Soft}) and (\ref{grandR}))
\STATE ${\bf f} \Leftarrow \GRD \UL\{{\bf f}\}$ (using (\ref{GRD}) and (\ref{grandU}))
\ENDWHILE
\STATE $f_{\TVL} \Leftarrow \UL\{{\bf f}\}$ (using (\ref{grandU}))
\STATE Re-normalize $f_{\TVL}$ (optional) 
\end{algorithmic}
\end{algorithm}

\section{Technical Remarks}
\subsection{MATLAB implementation}
For the sake of reproducibility of the results of this paper, some principal routines needed for implementation of the proposed method are detailed next. In particular, in this subsection, we provide examples of MATLAB$\textregistered$ codes for computation of operators $\GRD$, $\DIV$, $\Uu_L$, $\Aa_L$, $\Aa_L^\star$, $\Rr_L$, and the update equation (\ref{grandAlg}). (Note that the codes below have been optimized for clarity rather than for speed; substantial ``speed-ups'' are hence possible.) To this end, the variables {\tt u} and {\tt v} will be used to denote generic elements of $\Rnm$ and $\RRL$, respectively, with {\tt u} handled as an $\tt N\times M$ array, and {\tt v} handled as an $\tt N\times M\times 2*L$ array. 

Using the above notations, the MDG operator $\GRD$ can be computed by means of the m-function {\tt MDG} which is given below. Note that the function computes the MDG using periodic boundary conditions.
\begin{verbatim}
function [v] = MDG(u,L)
theta = (pi/2/L)*(0:L-1);
alpha=cos(theta);
beta=sin(theta);
[N,M]=size(u);
ux=u-u([N,1:N-1],:);
uy=u-u(:,[M,1:M-1]);
v=zeros(N,M,2*L);
for k=1:L,
    v(:,:,2*k-1)=alpha(k)*ux+beta(k)*uy;
    v(:,:,2*k)=alpha(k)*uy-beta(k)*ux;
end 
\end{verbatim}
The multidirectional divergence (MDD) operator $\DIV$ corresponding to $\GRD$ above can be implemented using the m-function {\tt MDD}.
\begin{verbatim}
function [u] = MDD(v,L) 
theta = (pi/2/L)*(0:L-1);
alpha=cos(theta);
beta=sin(theta);
[N,M,K]=size(v);
[ux,uy]=deal(zeros(N,M));
for k=1:L,
    ux=ux+(alpha(k)*v(:,:,2*k-1)-beta(k)*v(:,:,2*k));
    uy=uy+(beta(k)*v(:,:,2*k-1)+alpha(k)*v(:,:,2*k));
end
u=(ux([2:N,1],:)-ux)+(uy(:,[2:M,1])-uy);
\end{verbatim}

Having available the {\tt MDG} and {\tt MDD} functions, the computation of the remaining operators is straightforward. In particular, given the $\tt N\times M$ matrices {\tt H}  and {\tt Wi} of the transfer functions of the convolution and integration operators, respectively, and defining ${\tt A = H.*Wi}$, the operator $\Aa_L$ can now be computed using the m-function {\tt operator\_A} as given below.

\begin{verbatim}
function [u] = operator_A(v,A,L)
u=(1/L)*real(ifft2(fft2(MDD(v)).*A)));
\end{verbatim}
It is important to note that the integration operator $\Uu_L$ can be implemented using the same m-function {\tt operator\_A} with the substitution of {\tt Wi} for {\tt A}. Finally, the m-function {\tt operator\_A\_star} can be used to compute the adjoint operator $\Aa_L^\ast$
\begin{verbatim}
function [v] = operator_A_star(u,A,L)
v=(-1/L)*MDG(real(ifft2(fft2(u).*conj(A))));
\end{verbatim}
while the composite operator $\Rr_L$ can be implemented by the {\tt operator\_R} function defined as
\begin{verbatim}
function [v] = operator_R(v,A,L)
A2=A.*conj(A);
v=(-1/L^2)*MDG(real(ifft2(fft2(MDD(v)).*A2)));
\end{verbatim}

Using the above functions, the proposed method for TV-based image restoration can be implemented as a  series of updates performed according to (\ref{grandAlg}). In particular, given the parameters {\tt c} and {\tt tau} defined by lines 3 and 4 of Algorithm 2, respectively, and denoting by {\tt g} and {\tt f} the data image and its corresponding reconstruction, a total of {\tt K} updates can be performed using the following simple code\footnote{The {\tt wthresh} function is part of Wavelet Toolbox$^{\rm TM}$ of MATLAB\textregistered.}. 
\begin{verbatim}
b=operator_A_star(g,L);
v=MDG(g,L);                          %initialization
for k=1:K,
    v=wthresh(v+(1/c)*(b-operator_R(v,L)),'s',tau);
    v=MDG(operator_A(v,Wi,L),L);
end
f=operator_A(v,Wi,L);
f=(255/range(f(:)))*(f-min(f(:)));   %re-normalization
\end{verbatim}
Note that the last line of the above code forces the range of the recovered image {\tt f} to fit the interval $[0, 255]$, which by no means suggests it to be the only normalization scheme allowed.

\subsection{Possible ways to improve the rate of convergence}\label{speedup}
The computation of the multidirectional gradient and divergence operators can be performed with linear complexity as suggested by m-functions {\tt MDG} and {\tt MDD} above. Consequently, the computational cost of each update step of Algorithm 2 is determined by the cost of 2-D FFT, and hence it has a complexity of $\mathcal{O}\left( N M \log(N M) \right)$. Although such a complexity can be considered as standard for many iterative methods of image deconvolution, it is important to specify ways to reduce the total number of iterations performed by the algorithm, thereby minimizing its {\em overall} computational cost. To this end, it is first noted that the magnitude of the update term $(1/c) \left( {\bf b}_L - \Rr_L \{ {\bf f}^{(t)} \} \right)$ and the threshold $\tau = (\lambda d_L)/c$ in (\ref{grandAlg}) are inversely proportional to the value of $c$. Therefore, the larger the value of $c$, the more significant is the change in ${\bf f}^{(t)}$ caused by each iteration. Thus, to maximize the effect of the update (\ref{grandAlg}), the value of $c$ should be kept minimal.

The requirement to minimize $c$ appears to be at variance with the definition of its lower bound in (\ref{Lbound}), which suggests that $c$ grows as $\max\{N,M\}$ increases. This bound, however, should be considered as {\em uniform} in the sense of its being suitable for {\em all} iterations of the algorithm. It is known, on the other hand, that the method of iterative shrinkage belongs to the family of MM algorithms, in which a reduction in the value of an original cost functional is achieved through minimization of its {\em local} majorizers~\cite{Hunter04}. From this perspective, choosing $c$ in accordance with (\ref{Lbound}) provides an {\it a priori} guarantee on all the majorizers to be convex. Alternatively, at a given iteration $t$, to get a reduction in the value of the cost functional in (\ref{grandTV}), it is sufficient for $c$ to satisfy
\begin{equation}\label{Lcond}
c_t \left\| {\bf f}^{(t+1)} - {\bf f}^{(t)} \right\|_{\RRL}^2 \geq \left\| \Aa_L\{{\bf f}^{(t+1)} - {\bf f}^{(t)}\} \right\|_{\RRL}^2,
\end{equation}
where the subscript $t$ has been added to $c$ to express its dependency on the specific iteration. Consequently, one can ``fine-tune" the value of $c$ based on (\ref{Lcond}) by means of a simple back-tracking procedure as exemplified by Algorithm 3. Note that, in this algorithm, the value of $c$ is decreased exponentially (starting from an initial value obeying (\ref{Lbound})) by consecutively multiplying $c$ by a reduction factor $\mu \in (0,1)$.  

\begin{algorithm}
\caption{Local adjustment of the value of $c$ via back-tracking} 
\begin{algorithmic}[1]
\STATE $c \Leftarrow (1/L) \left[ 2 - 2 \cos \left( 2\pi \slash \max\{N,M\} \right) \right]^{-1} + \epsilon$ (for some $\epsilon > 0$)
\STATE $\tau \Leftarrow (\lambda \, d_L) \slash c$
\STATE ${\rm temp} \Leftarrow \mathcal{S}_\tau \left\{ {\bf f} + \left( {\bf b}_L - c^{-1} \Rr_L\{{\bf f}\} \right) \right\}$
\STATE ${\bf r} \Leftarrow {\rm temp} - {\bf f}$
\WHILE {$c \, \| {\bf r} \|_{\RRL}^2 \geq \left\langle \Rr_L\{{\bf r}\}, {\bf r} \right\rangle_{\RRL} $}
\STATE $c \Leftarrow \mu \, c$ (for some $\mu \in (0,1)$)
\STATE $\tau \Leftarrow (\lambda \, d_L) \slash c$
\STATE ${\rm temp} \Leftarrow \mathcal{S}_\tau \left\{ {\bf f} + \left( {\bf b}_L - c^{-1} \Rr_L\{{\bf f}\} \right) \right\}$
\STATE ${\bf r} \Leftarrow {\rm temp} - {\bf f}$
\ENDWHILE
\STATE ${\bf f} \Leftarrow \GRD \UL\{{\rm temp}\}$
\end{algorithmic}
\end{algorithm}

\begin{figure}[top]
\centering
\includegraphics[width=4in]{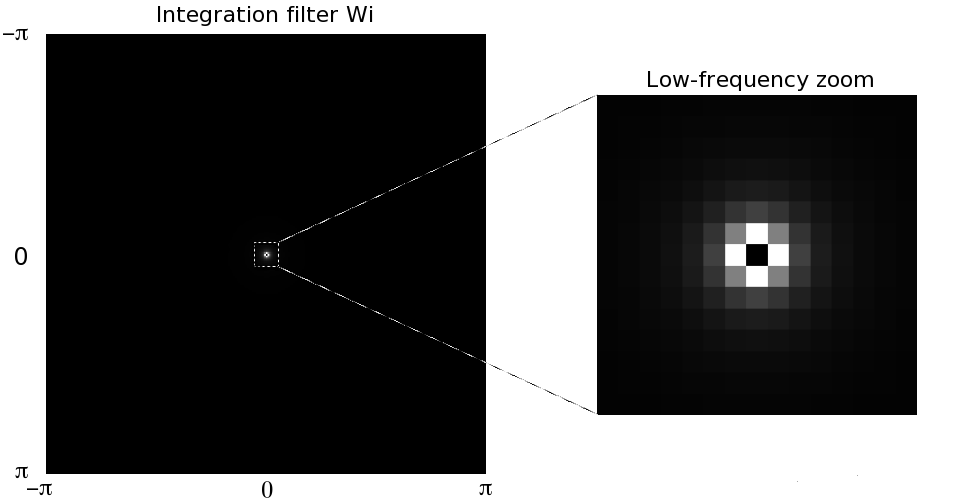}
\caption{The frequency response of the integration filter $W_i$ for the case of $N=M=256$. }
\label{F2}
\end{figure}

An alternative way to minimize the value of $c$ is to reduce the norm $\|\Aa_L \Aa_L^\ast\|$ through a sort of preconditioning. The latter can be defined by first noting that $\|\Aa_L \Aa_L^\ast\|$ is determined by the maximum value of $W_i\cdot |H|^2$. Moreover, since in most of the practically important cases, the frequency response $H$ can be normalized to satisfy $0 \leq (|H|^2)_{n,m} \leq 1$, it is mainly the values of $W_i$ which dominate the maximum of $W_i\cdot |H|^2$. Being an integration filter in nature, $W_i$ tends to amplify the lower frequencies. As a matter of fact, the values of $W_i$ are sharply peaked in a neighborhood of the zero frequency, as shown by Fig.~\ref{F2}. The figure visualizes $W_i$ as a gray-scale image for the case of $N=M=256$. One can see that the overwhelming portion of the values of $W_i$ are relatively small (as indicated by the black color), being equal to approximately 0.125. On the other hand, the values of the white pixels (which are few and hardly visible without a proper zoom) are close to 1160.1, which causes the norm $\|\Aa_L \Aa_L^\ast\|$ to have a relatively large value of 553.38, as computed according to (\ref{maxWi}) and (\ref{Norms}) for $L=3$. Consequently, the large values of $\|\Aa_L \Aa_L^\ast\|$ are, in fact, due to a negligibly small number of high-amplitude values of $W_i$ located around the DC.

The property of $|H|^2$ of being ``flat" around the zero frequency implies that its values cannot ``counterbalance" the high amplitudes of  $W_i$. On the other hand, one can {\it pre}-convolve the data image $g$ with another {\em auxiliary} filter whose DFT $H_c$ is designed so that the product $W_i \cdot |H_c|^2$ has a smaller maximum value than that of $W_i$. In this case, the resulting operator norm $\|\Aa_L \Aa_L^\ast\|$ will be defined by the maximum value of $W_i \cdot |H_c|^2 \cdot |H|^2$, which can be set to be reasonably small. It goes without saying that such a preconditioning changes neither the format of the restoration problem nor of its solution except for the obvious need to replace the original ``blur" $H$ by a new one, i.e. $H \cdot H_c$.

Additional techniques for further speeding-up the convergence of (\ref{grandAlg}) can be borrowed from the  field of projection methods of convex optimization~\cite{Birgin00}. In particular, let ${\bf f}^{(t)}$ and ${\bf f}^{(t+1)}$ be two successive outcomes of the shrinkage procedure (\ref{grandAlg}). Then, the next estimate of ${\bf f}$ can be found as a minimizer of the $\TVL$-functional along the {\em direction} ${\bf d}^{(t+1)} = {\bf f}^{(t+1)}-{\bf f}^{(t)}$. Note that, provided both ${\bf f}^{(t)}$ and ${\bf f}^{(t+1)}$ are in $\VV_L$, all solutions of the form ${\bf f}^{(t)} + \alpha \, {\bf d}^{(t+1)}$ (with $\alpha \in \mathbb{R}$) belong to $\VV_L$ as well, and therefore the above line search can be performed without using intermediate projection steps. It was demonstrated in~\cite{Elad06}, that augmenting the shrinkage operation by the line search can result in substantial increase in the rate of convergence of iterative shrinkage.

Finally, we note that the iterative shrinkage (\ref{grandAlg}) belongs to the family of so-called {\em one}-step shrinkage schemes. It was recently shown in~\cite{Dias07}, that the latter can be extended to {\em two}-stage shrinkage schemes, which have considerably higher rate of convergence and, therefore, provide additional means to further speed up the implementation of the proposed method for TV-based image restoration. 

\section{Results}
\subsection{Reference methods}\label{methods}
In this section, the theoretical results derived in the preceding sections of this paper are supported by a number of experimental results. In particular, we first show that image restoration by means of the proposed TV-based iterative shrinkage (TVIS) method with ${\rm TV} = \TVL$ and $L=3$ provides restoration results virtually indistinguishable from the results obtained using alternative methods of solving (\ref{TVgen}) with ${\rm TV} = \TVi$. Additionally, it will be shown that, in the case of strong convolutional blurs, the TVIS method can provide more valuable restoration results as compared with the method of~\cite{Figueiredo03}, which will be referred below to as the {\em sparse wavelet iterative shrinkage} (SWIS) method. Some minimal details about the references methods are given next for the sake of presentational completeness.
\subsubsection{The method of lagged diffusivity} This method (which seems to have been first proposed by  C. Vogel in~\cite{Vogel96}) is known to be one of the standard approaches to the solution of (\ref{TVgen}) with ${\rm TV}=\TVi$. The method is based on the first-order optimality condition for $E(f)$ in (\ref{TVgen}), which is given by 
\begin{equation}
\Hh^\ast\left\{ \Hh\{f\} - g \right\}  - \lambda \, {\rm div} \left( \frac{\nabla f}{ \| \nabla f\|} \right) = 0,
\end{equation}
where $\| \nabla f \| = \sqrt{ f_x^2 + f_y^2 }$. Consequently, the global minimizer of $E(f)$ is found as a stationary point of a sequence of solutions to the following system of equations
\begin{equation}\label{LD}
\Hh^\ast\left\{ \Hh\{f^{(t+1)}\}\right\}  - \lambda \, {\rm div} \left( \frac{\nabla f^{(t+1)}}{ \| \nabla f^{(t)}\|} \right) = \Hh^\ast\left\{ g \right\}
\end{equation}
solved w.r.t. $f^{(t+1)}$. Note that in (\ref{LD}), the result of previous iteration $f^{(t)}$ is considered to be constant (``frozen"), which makes (\ref{LD}) be a linear operator equation that can be solved iteratively by means of, e.g., conjugate gradient algorithm. It is also worthwhile noting that the method of~\cite{Vogel96} was recently rediscovered in~\cite{Bioucas-Dias06}, where the same iterative procedure (\ref{LD}) is derived using the majorization-minimization (MM) technique. What appears to be omitted in~\cite{Bioucas-Dias06}, however, is mentioning the fact that a stable implementation of (\ref{LD}) requires replacing the absolute value $\| \nabla f^{(t)} \|$ by its {\em strictly positive} approximation $\| \nabla f^{(t)} \| \approx \sqrt{(f_x^{(t+1)})^2 + (f_y^{(t+1)})^2 + \epsilon}$, for some $0 < \epsilon \ll 1$. Thus, strictly speaking, as long as $\epsilon > 0$, the global minimizer of $E(f)$ in (\ref{TVgen}) and the stationary point of (\ref{LD}) cannot be guaranteed to be identical, in general. It is possible, however, to converge to a close vicinity of the true global minimizer using a smooth relaxation procedure in which a stationary point of (\ref{LD}) is found for some value of $\epsilon>0$, followed by decreasing $\epsilon$ (e.g., $\epsilon \Leftarrow \epsilon/2$) and, subsequently, computing a new stationary point of (\ref{LD}) for the new $\epsilon$, while using the previous stationary point as an initialization. In the present paper, the above relaxation ``cycles" were performed to reduce the value of $\epsilon$ from $10^{-2}$ to $10^{-6}$.

Finally, we note that both the method of lagged diffusivity~\cite{Vogel96} and the proposed TVIS method require a definition of the regularization parameter $\lambda$. An optimal value of $\lambda$ can be elicited based on the theory of Bayesian estimation, according to which the TV-based image restoration assumes $\left( \| \nabla f \| \right)_{n,m}$ to be {\it i.i.d.} Laplacian, namely 
\begin{equation}\label{Laplace}
\left( \| \nabla f \| \right)_{n,m} \sim \frac{1}{2\beta} \exp \left\{ - \frac{\left( \| \nabla f \| \right)_{n,m}}{\beta}  \right\},
\end{equation}
where $\beta > 0$ is a scale parameter of the distribution which can be estimated as $\beta \approx \sqrt{0.5 \, \sigma_{\|\nabla f\|}^2}$, with $\sigma_{\|\nabla f\|}^2$ being the sample variance of $\| \nabla f\|$. Therefore, if the variance of the additive noise in (\ref{mdl}) is equal to $\sigma^2$, then from the viewpoint of MAP estimation, the optimal $\lambda$ should be set to be equal to $\lambda = \sigma ^2/\beta$, and this is how it was done in the present study.
\subsubsection{Sparse wavelet iterative shrinkage (SWIS)} Let $\{ \psi_k \}_{k \in \Gamma}$ be a dense set of wavelet functions in $\Rnm$, and $\Psi$ be their associated {\em synthesis} operator defined as
\begin{equation}\label{PSI}
\Psi: \ell_2(\Gamma) \leftarrow \Rnm: x \mapsto \sum_{k \in \Gamma} x_k \psi_k.
\end{equation}
In the present study, $\Psi$ is defined to correspond to a stationary separable wavelet transform, which implies that the set $\{ \psi_k \}_{k \in \Gamma}$ is overcomplete, and hence there is no unique way to represent an arbitrary $f\in\Rnm$ in terms of $\psi_k$. However, if $f$ can be {\em sparsely} represented by such a wavelet {\em dictionary}, then a useful approximation of $f$ in (\ref{mdl}) can be computed as
\begin{equation}\label{SWAS}
f \approx \argmin_{f \in \Rnm} \left\{ \frac{1}{2} \left\|  \Hh\left\{ \Psi\{x\} \right\} - g \right\|_2^2 + \lambda \, \| x \|_1 \right\},
\end{equation}
where $\lambda > 0$ is a regularization parameter analogous to that in (\ref{TVgen}). Moreover, it was proven in\cite{Daubechies03, Figueiredo03} that the above minimization problem can be solved via iterative shrinkage performed according to
\begin{equation}
x^{(t+1)} = \mathcal{S}_{\lambda/c} \left\{ x^{(t)} + \frac{1}{c} \Aa^\ast \left\{ g - \Aa \{ x^{(t)} \} \right\} \right\},
\end{equation}
where $\Aa\{\cdot\}$ is the composition of $\Hh$ and $\Psi$, $\Aa^\ast$ is the adjoint of $\Aa$, $\mathcal{S}_\tau$ is given by (\ref{Soft}), and the constant $c$ satisfies $c > \| \Aa \Aa^\ast \|$.

Note that, from the perspective of Bayesian estimation, the representation coefficients $x$ in (\ref{SWAS}) are assumed to be {\it i.i.d.} Laplacian, which justifies setting $\lambda$ to be equal to $\sigma^2/ \beta$, with $\beta$ being the scale parameter of the distribution of $x$. To estimate $\beta$, all the test images used in this study were first processed by the Basis Pursuit algorithm~\cite{Chen01} to find their corresponding sparse representations. Subsequently, the sample variances $\sigma_x^2$ of the sparse coefficients were estimated, followed by estimating the respective $\beta$ as $\sqrt{0.5 \,\sigma_x^2}$.

\subsection{TVIS versus the method of lagged diffusivity}
Practical implementation of the method of lagged diffusivity~\cite{Vogel96} requires one to optimally preset a number of ``internal" algorithmic parameters such as a convergence rule for the linear solver, the rate of relaxation for $\epsilon$, etc. Besides the somewhat esoteric nature of this preset, the latter also appears to be problem-dependent, which implies the necessity to adjust the algorithmic parameters for different data sets. The proposed TVIS algorithm, on the other hand, has a very simple structure which is devoid of any parameters that could be potentially dependent on data. Moreover, in Section~\ref{speedup}, a number of means to increase the rate of convergence of the TVIS algorithm were proposed, which could be useful, for example, in the case of processing large sets of imagery data. We reserve a thorough discussion of that and related matters for another occasion. Instead, in this section, we focus on finding the minimal number of multidirectional gradients $L$ in (\ref{TVapprox}) for which the restoration results obtained with the reference method of~\cite{Vogel96} and TVIS can be considered as comparable.

\begin{figure}[top]
\centering
\includegraphics[width=5in]{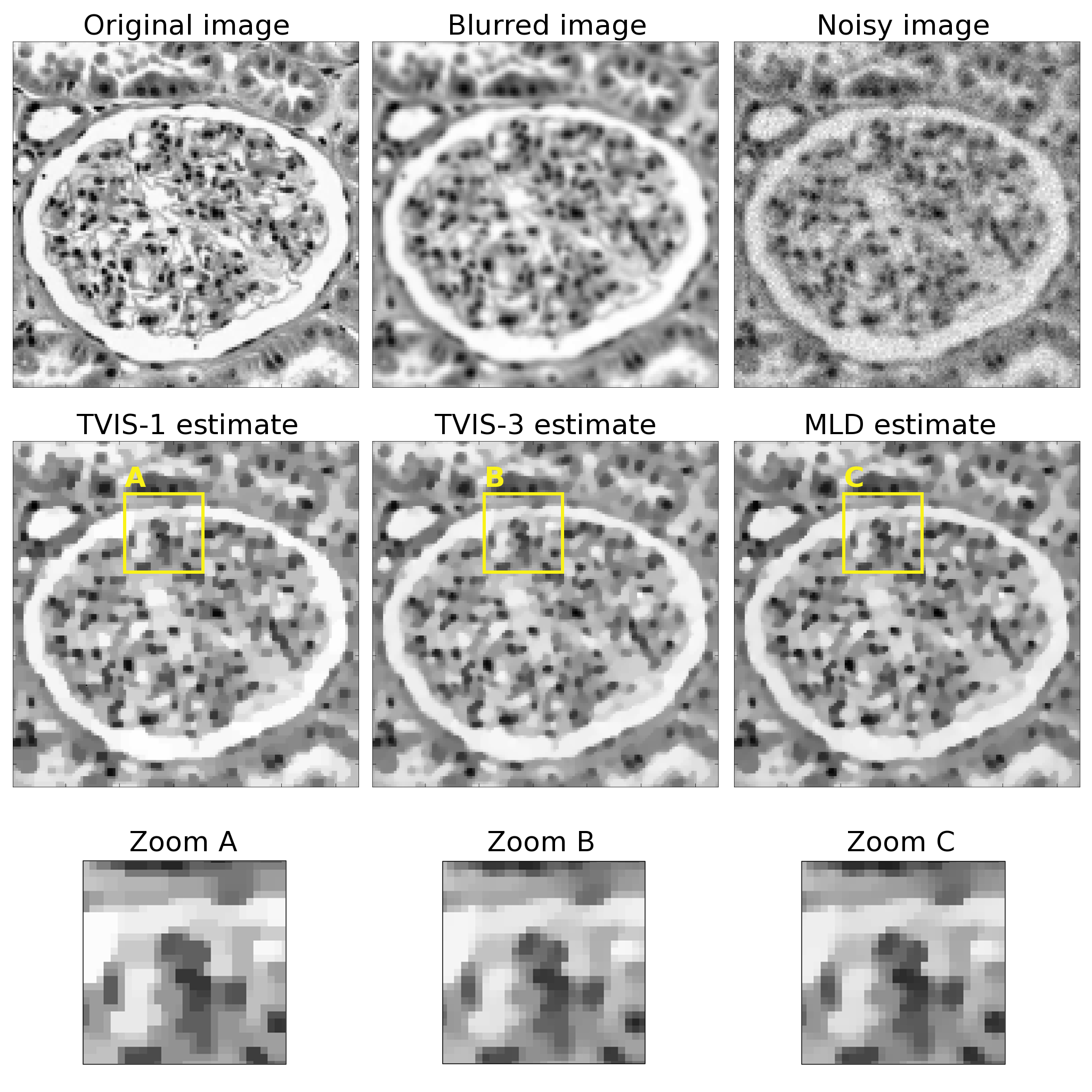}
\caption{(Upper row of subplots) Original image of glomerulus, its blurred and noisy versions (PSNR = 21.3 dB); (Middle row of subplots) Results of image restoration by (from left to right) TVIS-1, TVIS-3, and MLD; (Lower row of subplots) Zoomed segments of the estimated images as indicated by the yellow rectangles.}
\label{F3}
\end{figure}

\begin{figure}[top]
\centering
\includegraphics[width=5in]{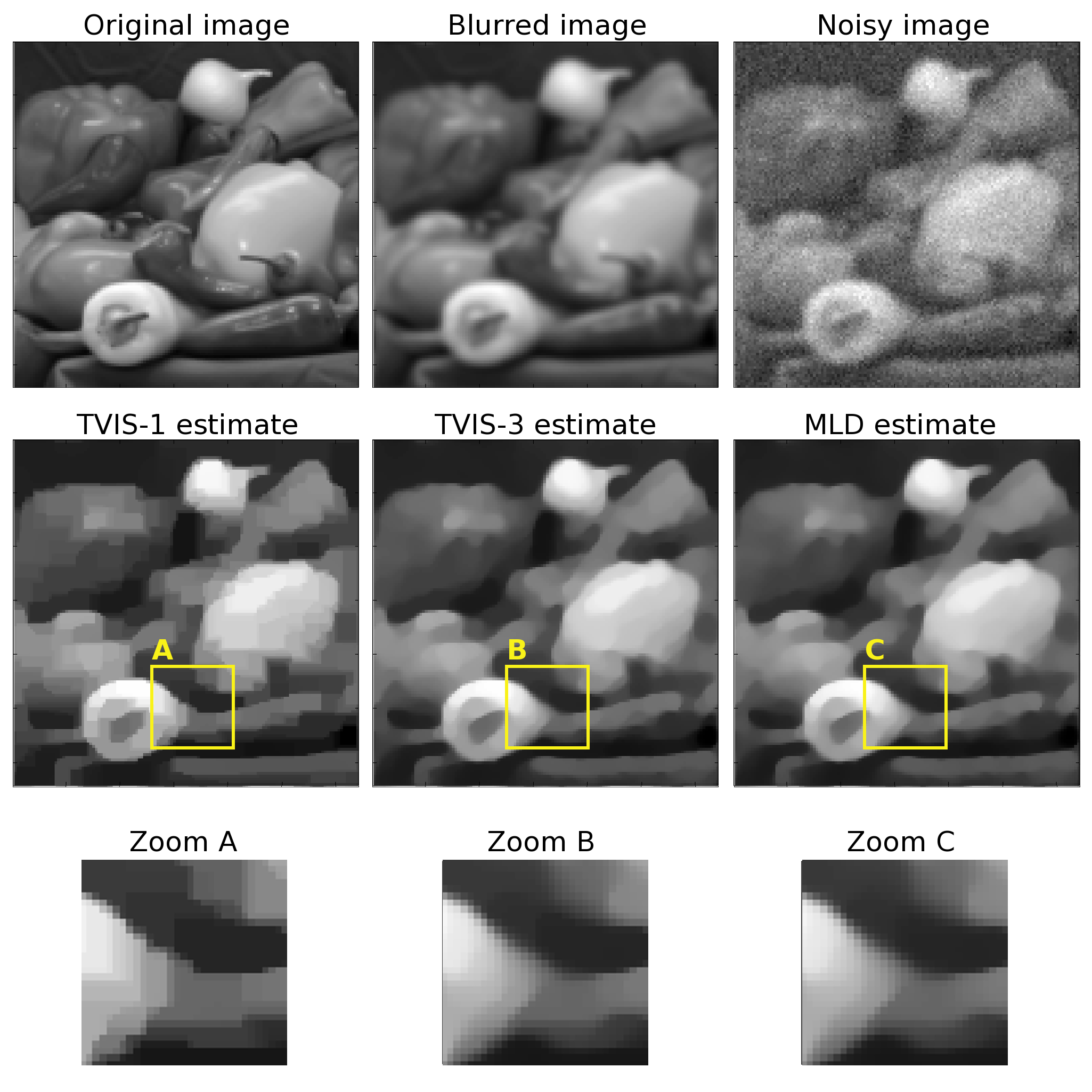}
\caption{(Upper row of subplots) Original image of peppers, its blurred and noisy versions (PSNR = 24.5 dB); (Middle row of subplots) Results of image restoration by (from left to right) TVIS-1, TVIS-3, and MLD; (Lower row of subplots) Zoomed segments of the estimated images as indicated by the yellow rectangles.}
\label{F4}
\end{figure}

In the examples below, two different values of the parameter $L$ of the TVIS algorithm are used, namely $L=1$ and $L=3$. Note that, in the case when $L=1$, TVIS performs {\em anisotropic} TV-based image restoration, whereas it approximates the solution of the {\em isotropic} TV-based image restoration problem when $L=3$. (The above two settings will be referred below to as TVIS-1 and TVIS-3, respectively). The legitimacy and usefulness of the above approximation are demonstrated through our first example in Fig.~\ref{F3}. The upper row of subplots in the figure shows an image of glomerulus, its blurred version, as well as the noisy image obtained by contaminating the blurred image with a white Gaussian noise. The blurring artifact was modeled by convolving the test image with a Gaussian kernel of standard deviation 0.8. It should be noted that, in this case, the condition number of the corresponding convolution operator $\Hh$ (which can be computed as the ratio of the maximum and minimum values of the DFT of the Gaussian kernel) is equal to 138.4, which is relatively small. The blur artifact in Fig.~\ref{F3}, therefore, can be classified as mild. The noise contamination, on the other hand, is relatively strong, resulting in the peak signal-to-noise ratio (PSNR) of 21.3 dB. 

The results of TV-based restoration of the image of glomerulus by TVIS-1, TVIS-3, and the method of lagged diffusivity (MLD) are shown in the middle row of subplots in Fig.~\ref{F3}. For all these methods, the same regularization parameter $\lambda = 0.028$ was used, which was found based on the methodology detailed in Section~\ref{methods}. Interestingly enough, all the estimated images appear to be virtually indistinguishable. A closer examination, however, reveals that the anisotropic restoration by TVIS-1 has more ``squarish" details as compared to the restorations by TVIS-3 and MLD. (An example of such a behavior is demonstrated by the zoomed segments of the image estimates shown in the lower row of subplots in Fig.~\ref{F3}.) The restorations obtained with TVIS-3 and MLD, on the other hand, {\em are} very close to each other, with a relative error between them being less than 0.5\%. Moreover, both methods resulted in the same PSNR of 24.4 dB.

Despite the profound similarity between the restoration results in Fig.~\ref{F3}, the conclusion that $\TVa$- and $\TVi$-based image restorations are similar would be rather premature as proven by the example in Fig.~\ref{F4} (whose composition is analogous to that of Fig.~\ref{F3}). In this case, the standard image of peppers has been blurred by a Gaussian kernel of standard deviation 1.2, with the condition number of its associated $\Hh$ being equal to 326217.8. The addition of white Gaussian noises gave rise to the PSNR of 24.5 dB.

Analyzing the restoration results shown in the middle row of subplots in Fig.~\ref{F4}, one can hardly see any difference between the estimates provided by TVIS-3 and MLD. Both methods resulted in the same improvement of 4.3 dB in terms of the PSNR, which is natural considering the fact the relative error between the TVIS-3 and MLD restorations was found to be below 0.5\%. The anisotropic nature of TVIS-1, on the other hand, can now be clearly seen. This nature manifests itself in the irregular behavior of the image contours, which are smooth in the original scene. In this example, therefore, the restoration by TVIS-3 should be preferred over that by TVIS-1. 

\subsection{TVIS versus SWIS}
Based on the results of the preceding subsection one can conclude that, for the case of $L=3$, the estimations obtained with the TVIS algorithm are very close to what can be obtained using alternative methods for $\TVi$-based image restoration. In this subsection, therefore, only the performance of TVIS-3 is compared with that of the method of image deconvolution by sparse wavelet iterative shrinkage (SWIS)~\cite{Figueiredo03, Elad07}. To sparsely represent test images, stationary (aka non-decimated) separable wavelet transforms with three resolution levels were used. Note that, in this case, the number of wavelet coefficients is ten times higher than the number of image pixels, which suggests that the wavelet transform has the overcomplete factor of 10:1. The TVIS-3 algorithm, on the other hand, has smaller storage requirements, as it operates with a total of six partial derivatives, resulting in the overcomplete factor of 6:1.    

\begin{figure}[top]
\centering
\includegraphics[width=4.5in]{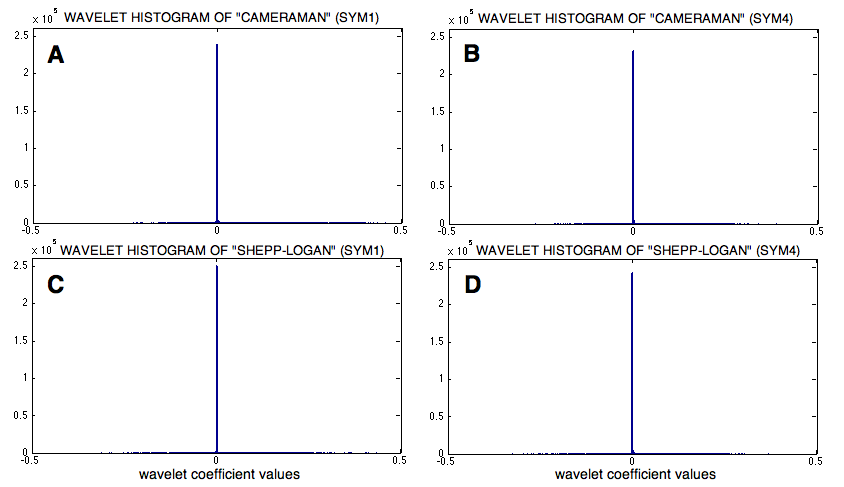}
\caption{(Subplot A) Histogram of the stationary wavelet coefficients of ``Cameraman" computed by BP with the ``Symlet-1" wavelet; (Subplot B) Histogram of the stationary wavelet coefficients of ``Cameraman" computed by BP with the ``Symlet-4" wavelet; (Subplot C) Histogram of the stationary wavelet coefficients of ``Shepp-Logan" computed by BP with the ``Symlet-1" wavelet; (Subplot D) Histogram of the stationary wavelet coefficients of ``Shepp-Logan" computed by BP with the ``Symlet-4" wavelet.}
\label{F7}
\end{figure}

\begin{figure}[top]
\centering
\includegraphics[width=5in]{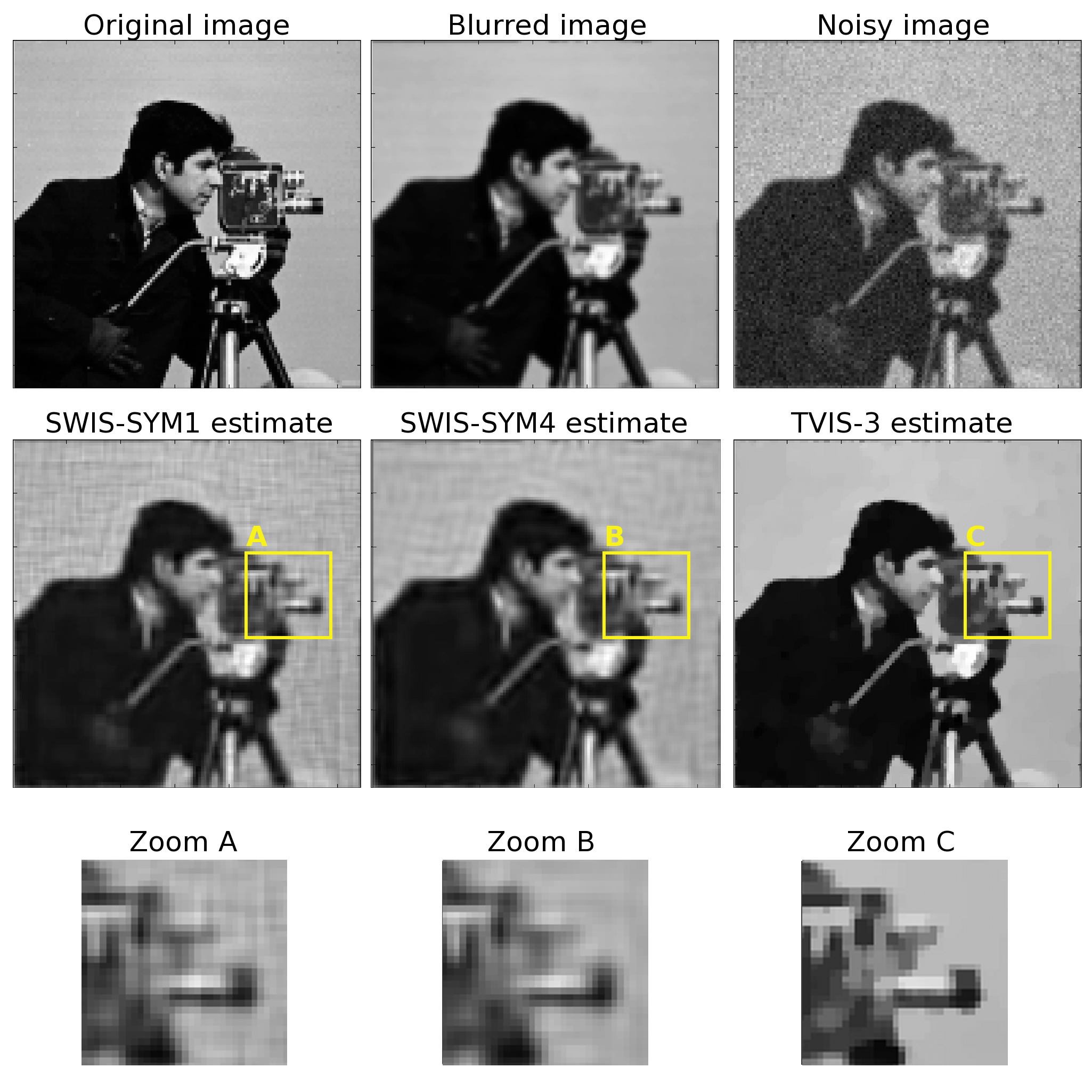}
\caption{(Upper row of subplots) Original image of "Cameraman", its blurred and noisy versions (PSNR = 22.4 dB); (Middle row of subplots) Results of image restoration by (from left to right) SWIS-SYM1, SWIS-SYM4, and TVIS-3; (Lower row of subplots) Zoomed segments of the estimated images as indicated by the yellow rectangles.}
\label{F5}
\end{figure}

\begin{figure}[top]
\centering
\includegraphics[width=5in]{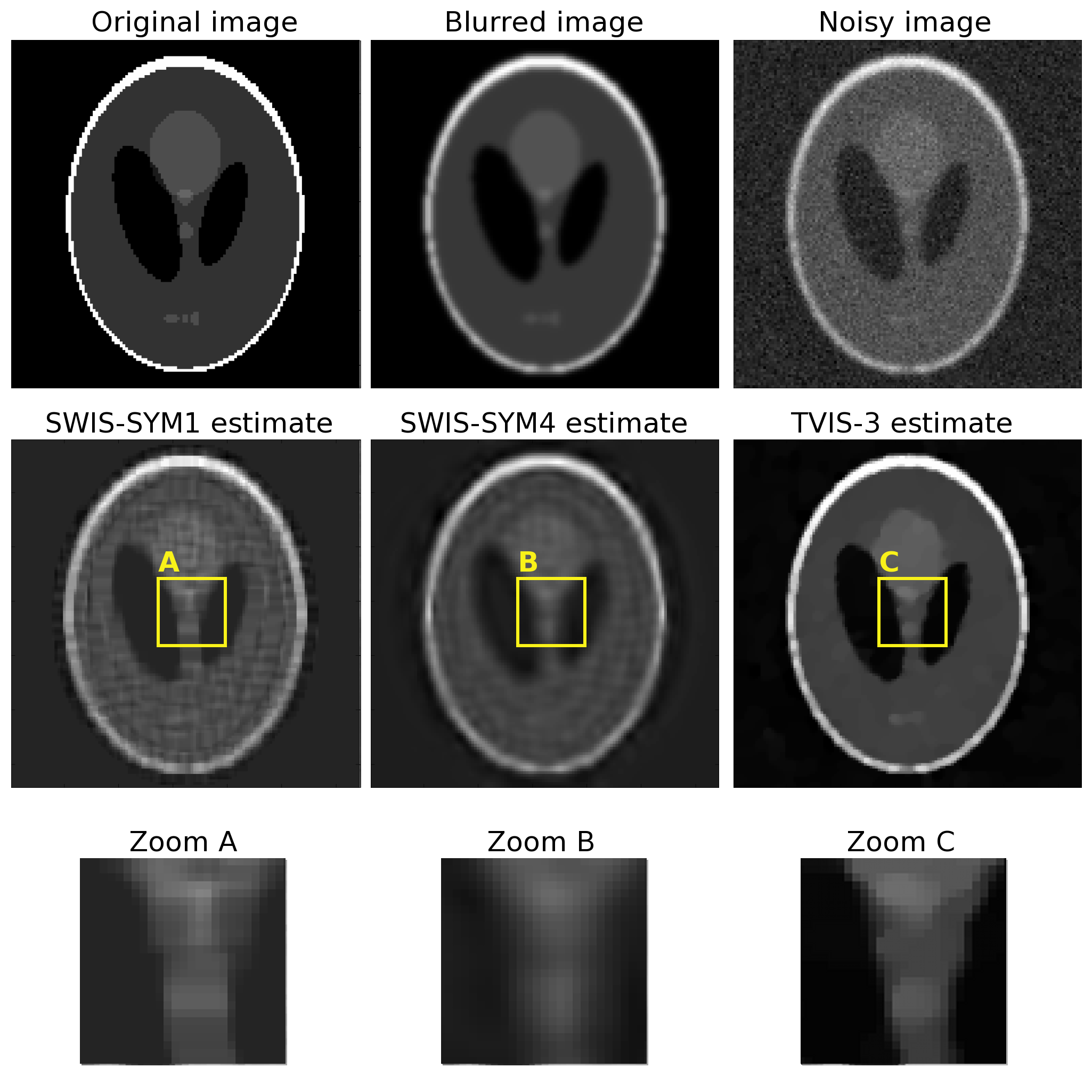}
\caption{(Upper row of subplots) Original image of "Shepp-Logan phantom", its blurred and noisy versions (PSNR = 19.0 dB); (Middle row of subplots) Results of image restoration by (from left to right) SWIS-SYM1, SWIS-SYM4, and TVIS-3; (Lower row of subplots) Zoomed segments of the estimated images as indicated by the yellow rectangles.}
\label{F6}
\end{figure}

At the heart of the SWIS method is the assumption that the image of interest can be sparsely represented in the domain of a wavelet transform. To verify this assumption, the basis pursuit (BP) algorithm~\cite{Chen01} was employed to find sparse representations of the standard ``Cameraman" and ``Shepp-Logan phantom" images in the domain of the wavelet transforms corresponding to the nearly-symmetric wavelet of I.~Daubechies having 1 and 4 vanishing moments\footnote{In the case of one vanishing moment, the wavelet is identical to the ``Haar" wavelet.}. (Since in the standard nomenclature, the above wavelet functions are referred to as Symlet1 and Symlet4, the corresponding SWIS algorithms will be referred below as SWIS-SYM1 and SWIS-SYM4, respectively). Subplot A and B of Fig.~\ref{F7} show the histograms of the wavelet coefficients of ``Cameraman" corresponding to Symlet1 and Symlet4, respectively, while the histograms of the wavelet coefficients of ``Shepp-Logan phantom" computed using the above wavelets are shown in Subplots C and D of the same figure. The profoundly super-Gaussain appearances of all the histograms suggests that both test images are indeed sparsely representable by the wavelet transforms in use.

The upper row of subplots of Fig.~\ref{F5} show the original image of ``Cameraman" as well as its blurred and noisy versions. In this example, the ``mild" Gaussian blur of standard deviation 0.8 was used to smooth the image, followed by addition of while Gaussian noise giving rise to the PSNR of 22.4 dB. Subsequently, the image was subjected to the restoration procedures of SWIS-SYM1, SWIS-SYM4, and TVIS-3 algorithms, whose results are depicted in the middle row of subplots in Fig.~\ref{F5}. In all the above cases, the regularization parameter $\lambda$ was computed according to the methodology discussed in Section~\ref{methods}. Specifically, for SWIS-SYM1, SWIS-SYM4, and TVIS-3, the optimal values of $\lambda$ were found to be equal to 0.1335, 0.1505, and 0.024, respectively.

Analyzing Fig.~\ref{F5}, one can see that the best restoration results in terms of contrast improvement and noise reduction are obtained by TVIS-3, with its PSNR equal to 26.8 dB. At the same time, both SWIS-SYM1 (PSNR = 22.8 dB) and SWIS-SYM4 (PSNR = 22.4 dB) suppress the additive noise, while being unable to surmount the effect of blur. It is also interesting to note that SWIS-SYM1 provides a ``sharper" reconstruction as compared with SWIS-SYM4, while the latter results in smoother estimation of the uniform background.

The inability of SWIS to overcome the effect of strong blurs is not occasional. This fact is further supported by Fig.~\ref{F6} which demonstrates the results of restoration for ``Shepp-Logan phantom". In this example, the (relatively strong) Gaussian blur of standard deviation 1.2 was used to smooth the original image, which together with noise contamination resulted in the PSNR of 19.0 dB. Analyzing Fig.~\ref{F6}, one can once again see that TVIS-3 is capable of accurately recovering the piecewise constant structure of ``Shepp-Logan phantom" (PSNR = 23.9 dB), while neither SWIS-SYM1 (PSNR = 20.8 dB) nor SWIS-SYM4 (PSNR = 20.9 dB) could de-noise and sharpen the image to a similar extent. Moreover, we see again that SWIS-SYM1 produces ``sharper" (yet noisier) restoration as compared to SWIS-SYM4. 

The reason why SWIS cannot reduce strong blurring artifacts does not lie in the nature of the particular solution method used, but is intrinsic in the way the regularization is performed. Specifically, for the sake of illustration, let us assume that $\Hh$ in (\ref{mdl}) is a low-pass {\em half-band} filter and no noise has been observed in the measurement of $\Hh\{f\}$, i.e. $e = 0$. In this case, there is an infinite number of possible solutions to $\Hh\{f\} = g$, all of which are different over the null space of $\Hh$ containing high-frequency signals whose spectra vanish below the cut-off of $\pi/2$. Among these candidate solutions, we are interested in specific two, namely $f$ and $g$ (note that $\Hh\{g\} = g$, since $\Hh$ is nilpotent). The sparse regularization via $\ell_1$-norm minimization can ``prefer" $f$ over $g$ only if $\|f\|_1 < \|g\|_1$. This is, however, not true in general. For example, in the case of Shannon multiresolution~\cite[Ch.VII]{Mallat98}, all the wavelet coefficients of $f$ and $g$ would be identical, except for the highest resolution level, where the coefficients of $g$ would necessarily vanish as opposed to those of $f$. Obviously, in this case, $\|f\|_1\geq \|g\|_1$ holds for any pair $(f, g)$, and, therefore, the minimization of the $\ell_1$-norm would result in $g$ as a final solution, not $f$.

Even though ideal blurs are rare in practice (as well as the use of Shannon wavelets), the above considerations remain relevant in the case of relatively strong blurs and regular wavelet analysis. This seems to be the reason for which the examples in~\cite{Figueiredo03} and~\cite{Elad07} used relatively weak blurs and irregular Haar wavelets. In the case of heavier blurring artifacts, however, the assumption on the sparsity of wavelet coefficients (as measured by the $\ell_1$-norm) does not seem to be sufficiently strong to guarantee a successful image restoration as exemplified by Fig.~\ref{F5} and Fig.~\ref{F6}. Under such conditions, performing TV-based image restoration (by means of, e.g., TVIS) should be considered as a more effective alternative.

\section{Discussion and Conclusions}
Among the existing methodologies for restoration of digital images, the TV-based method of~\cite{Rudin92} is known to be one of the most fundamental techniques. In this method, the instability of image restoration caused by the property of $\Hh$ in (\ref{mdl}) being a poorly conditioned operator is overcome by using a priori information. The latter is introduced in the form of a requirement on the recovered image to have a small total-variation norm. In general, the solution of such a regularized problem depends on the definition of the TV-norm, which can be either anisotropic or isotropic. Thus, for some images, both definitions can result in very close estimations, while for others, the results may differ dramatically. In this respect, the proposed TVIS method is advantageous, for it allows one to switch between the above two settings by merely changing the value of $L$ in (\ref{TVapprox}). Moreover, it was demonstrated both conceptually and experimentally that, for $L=3$, the TVIS algorithm provides estimation results which are virtually indistinguishable from the results obtained by alternative methods of $\TVi$-based image restoration.

Yet another useful feature of the TVIS algorithm consists in its particularly simple structure, which requites only a recursive application of linear filtering and soft-thresholding. Furthermore, as opposed to many alternative methods of TV-based image restoration, TVIS does not require presetting any algorithmic parameters that could potentially depend on the properties of imagery data.

The rate of convergence of the TVIS algorithm is defined by the image size as well as by the value of $L$ according to (\ref{Lbound}). In Section~\ref{speedup}, however, a number of practical ways were detailed, using which one can substantially reduce the total number of iterations required by the algorithm. For the reason of limited space, a comparative analysis of these ``speedup" schemes has not been included in the present paper, and it will be published elsewhere. However, even in its current form (see Algorithm 2), the proposed method has a complexity compared to that of the sparse wavelet iterative shrinkage in~\cite{Figueiredo03, Daubechies03, Elad07}. Yet, as opposed to SWIS, TVIS is capable of reliably recovering the images corrupted by relatively strong noises and blurring artifacts, as demonstrated by Fig.~\ref{F5} and Fig.~\ref{F6}.

\section{Acknowledgment}
The author would like to thank Elad Shaked for very fruitful and interesting discussions on the subject of TV-based image restoration.

\end{document}